%% file: main.tex
\documentclass{article}
\usepackage[final]{acl}  

\input{math_commands.tex}  

\usepackage{times}
\usepackage{latexsym}

\usepackage{amsmath}
\usepackage{xcolor}
\usepackage{cleveref}
\usepackage[normalem]{ulem}
\usepackage{xspace}
\usepackage{enumitem}
\usepackage{booktabs}
\usepackage{soul}
\usepackage{subcaption}
\usepackage{wrapfig}
\usepackage[T1]{fontenc}
\usepackage[utf8]{inputenc}
\usepackage{microtype}
\usepackage{inconsolata}
\usepackage{graphicx}
\usepackage{tcolorbox}
\tcbuselibrary{breakable}
\usepackage{fancyvrb}
\usepackage{float}
\usepackage{placeins}

\title{Self-Consistency from Only Two Samples:\\ CoT–PoT Ensembling for Efficient LLM Reasoning}
 
\author{
  Raman Saparkhan$^1$ \quad Majd Hawasly$^2$ \quad Md Rizwan Parvez$^2$ \quad Mohammad Raza$^2$ \\
  $^1$Carnegie Mellon University Qatar \quad $^2$Qatar Computing Research Institute \\
  \small\texttt{rsaparkh@andrew.cmu.edu} \quad \texttt{mhawasly@hbku.edu.qa} \quad \texttt{mparvez@hbku.edu.qa} \quad \texttt{mraza@hbku.edu.qa}
}

\begin{document}
\maketitle
\thispagestyle{plain}

\begin{abstract}
Self-consistency (SC) is a popular technique for improving the reasoning accuracy of large language models by aggregating multiple sampled outputs, but it comes at a high computational cost due to extensive sampling. We introduce a hybrid ensembling approach that leverages the complementary strengths of two distinct modes of reasoning: Chain-of-Thought (CoT) and Program-of-Thought (PoT). We describe a general framework for combining these two forms of reasoning in self-consistency, as well as particular strategies for both full sampling and early-stopping. We show that CoT-PoT ensembling not only improves overall accuracy, but also drastically reduces the number of samples required for SC by a factor of $9.3\times$. In particular, the majority of tasks ($78.6\%$) can be addressed with \emph{only two} samples, which has not been possible with any prior SC methods.
\end{abstract}

\input{introduction}
\input{method.tex}
\input{evaluation}

\input{selftraining}

\input{relatedwork}
\input{conclusion}
\input{limitations}

\bibliography{custom}

\appendix
\include{appendix}  

\end{document}

%% file: math_commands.tex
\usepackage{amsmath,amsfonts,bm}

\def\eqref#1{equation~\ref{#1}}

\def\1{\bm{1}}

\DeclareMathAlphabet{\mathsfit}{\encodingdefault}{\sfdefault}{m}{sl}
\SetMathAlphabet{\mathsfit}{bold}{\encodingdefault}{\sfdefault}{bx}{n}



%% file: introduction.tex
\section{Introduction}
\label{sec:intro}

Reliable reasoning remains an important challenge for large language models (LLMs), and the paradigm of self-consistency (SC) is a widely adopted approach for improving the robustness of LLM reasoning at inference time ~\citep{wang2022self}. In this ensembling method, multiple outputs representing different reasoning paths are generated by the same model for a given input problem and the final answer is selected as the most frequently occurring one among these samples. While this approach yields higher accuracy than a single inference, it also comes with significantly higher computational cost, as numerous inference calls to the LLM are required, e.g. 40 samples are common to reach the best accuracy levels. Various approaches have therefore been proposed to reduce the number of samples required in self-consistency ~\citep{adaptiveconsistency,escapeskyhighcost,difficultyadaptiveconsistency}. For instance, \emph{adaptive consistency} is an early-stopping technique where sampling is terminated if a confident majority is established early on~\citep{adaptiveconsistency}. While showing relative improvements in efficiency, such approaches still require many samples and yield at best comparable—and often lower—accuracy than full sampling. Hence, improving both the accuracy and efficiency of self-consistency is an important challenge, 
especially as inference-time scale-up is increasingly used to handle complex reasoning tasks with ever-larger models ~\citep{deepseekr1,openai_introducing_gpt5_2025}.

\begin{figure*}[t]  
  \centering
\includegraphics[width=\linewidth]{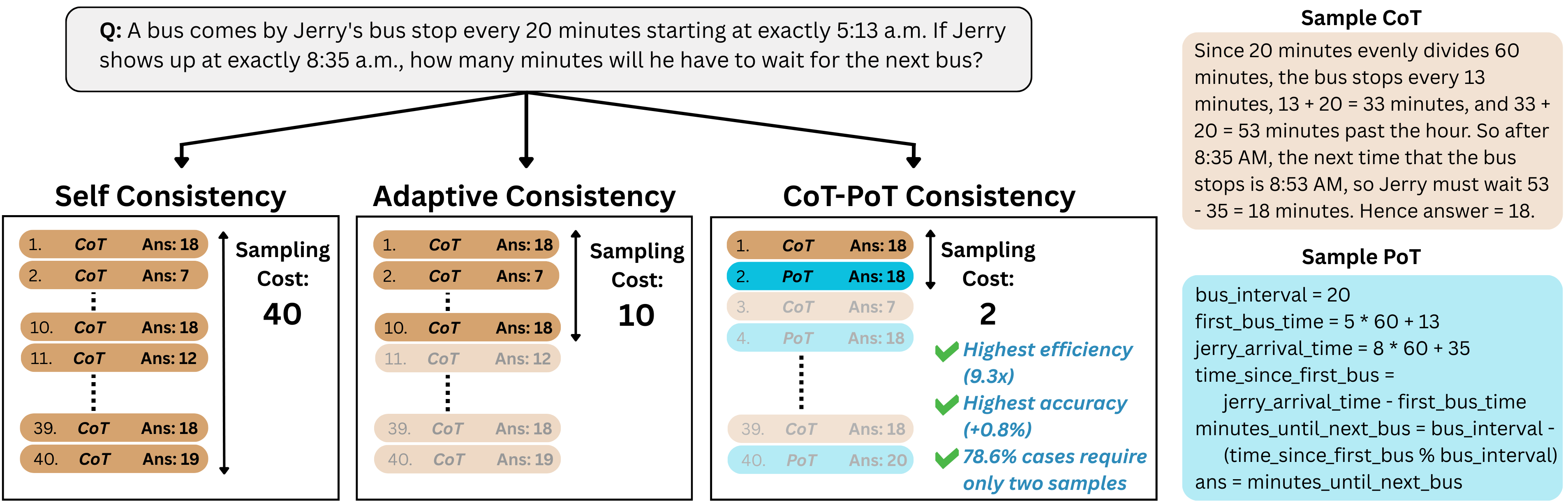}
  \caption{CoT-PoT consistency provides the highest accuracy, the highest efficiency and can solve most problems with only two samples, unlike any prior  self-consistency method.} 
  \label{fig:sc-vs-adapt-vs-cotpot}
\end{figure*}

In this work we address this challenge with a novel approach to self-consistency that combines different reasoning modalities. The core idea behind self-consistency is that if different ways of reasoning converge on the same final answer, then such consensus serves as a strong signal of correctness. Hence, what matters is the \emph{diversity} of the reasoning paths rather than just quantity. Existing SC techniques rely on high model temperatures to induce such diversity, but in practice we observe that this often yields reasoning traces that are very similar and may only have superficial syntactic variations in wording rather than substantial semantic differences. We address this issue with a new SC approach that is based on two fundamentally distinct modes of reasoning: chain-of-thought (\emph{CoT}) ~\citep{cotpaper} and program-of-thought  (\emph{PoT})~\citep{potpaper, palpaper}. CoT is a concrete form of reasoning in natural language where the model generates step-by-step inferences to explicitly construct a final answer. In contrast, PoT is a more abstract or symbolic form of reasoning where the model formulates the solution as a program that is executed to compute the final answer. While prior work has explored combining CoT and PoT reasoning for various purposes~\citep{mammoth,automaticmodelselection23,llmcascades}, the primary novelty of our work is in leveraging \emph{cross-modal agreement as a reliable early-stopping signal for self-consistency}, enabling drastic reductions in sampling cost that no prior approach has achieved.

Figure \ref{fig:sc-vs-adapt-vs-cotpot} illustrates this contrast with an example query about bus scheduling along with a sample CoT and PoT solution (right). The CoT solution first makes a simplifying inference that the bus always stops at the same number of minutes past the hour, and then does all arithmetic calculations explicitly to infer the requested waiting time. Many other CoT samples from temperature sampling may follow a similar pattern of reasoning with variations only in style or phrasing. In contrast, the PoT solution creates a symbolic representation of the whole scenario, representing actual times as minutes past midnight, formulating the answer as the total time since the first bus modulo the interval, and using program execution to perform all the arithmetic calculations. Due to the different reasoning approaches, the two modalities also exhibit different kinds of errors: for instance, CoTs may perform calculation errors (e.g. "53 - 35 = 8"), while PoTs may incorrectly express symbolic relationships (e.g. use a division "/" operator instead of modulo "\%"). Thus, the two modalities exhibit complementary strengths but have different error modes -- CoT is more concrete and flexible but it can suffer from imprecision and computational errors, while PoT provides stronger computational robustness but may make symbolic formulation errors. The agreement between the two modalities would therefore mean alignment between logical framing and computational aspects.

Our \textit{CoT-PoT ensembling} approach leverages the high diversity and complementary strengths of these two reasoning modalities to improve both the accuracy and efficiency of SC inference. Firstly, we explore cross-modal full sampling strategies that utilize the entire sampling budget, and show that aggregating over both CoT and PoT samples provides higher accuracy than either CoT-only or PoT-only self-consistency. Moreover, a key observation of this work is that agreement between CoT and PoT responses also provides a strong signal for \emph{early stopping} of sampling. Such agreement is highly informative  since the error modes of the two approaches have very low correlation, unlike near-identical samples in the same modality that may make similar errors. We formalize this notion by first formulating a general Bayesian model of cross-modal agreement, and then study specialized instantiations of this model that represent different early-stopping sampling strategies. We investigate both data-driven strategies, in which model parameters are learned from held-out data, and data-independent strategies based on extreme parameterizations.   

Figure \ref{fig:sc-vs-adapt-vs-cotpot} illustrates our most efficient early-stopping CoT-PoT method, where we alternately sample one CoT and one PoT solution until there is agreement between the two modalities. In this example scenario, we see that while the standard CoT-only self-consistency method requires the full budget of 40 samples, and the adaptive consistency early stopping method requires 10 samples to establish a statistical majority, our CoT-PoT method requires only 2 samples based on agreement between the initial CoT and PoT solutions. Over a range of diverse benchmarks and different LLMs, we show how our early-stopping CoT-PoT techniques achieve higher accuracy than any prior self-consistency approaches, while also providing the highest reduction in the number of samples, by a factor of $9.3\times$. Furthermore, another distinguishing aspect of our approach is that most tasks can be handled with \emph{only two} samples, which has not been possible with any prior self-consistency technique (the best early-stopping methods require at least four samples in any case). On average, our CoT-PoT method can handle $78.6\%$ of tasks using only two samples, with this number exceeding $90\%$ for some benchmarks and models. In particular, we find that more reasoning-intensive models such as DeepSeek R1 benefit more from such two-sample cross-modal consistency. This makes our method highly efficient for most problems, while still providing greater overall robustness over prior approaches. 

In summary, we make the following contributions in this work: (1) We investigate full sampling methods combining CoT and PoT modalities, and show that such methods provide higher accuracy than standard SC under the same sampling budget. (2) We develop a general Bayesian model of cross-modal agreement, and derive both data-driven and data-independent early stopping sampling strategies as instantiations of this model. (3) We provide an empirical evaluation over five diverse benchmarks and five mainstream LLMs as well as one small model. This shows how CoT-PoT ensembling provides the highest accuracy and efficiency over prior methods, solving most problems with only two samples.

%% file: method.tex
\section{CoT-PoT Ensembling}
\label{sec:method}

\newcommand{\anscot}{S^{\text{c}}}
\newcommand{\anspot}{S^{\text{p}}}
\newcommand{\freqcot}[1]{\text{F}_{\text{c}}(#1)}
\newcommand{\freqpot}[1]{\text{F}_{\text{p}}(#1)}
\newcommand{\allfreq}[1]{\freqcot{#1} + \freqpot{#1}}
\newcommand{\methodname}[1]{\textbf{\texttt{#1}}\xspace}

\newcommand{\selfconsistency}{\methodname{SC\textsubscript{CoT}}} 
\newcommand{\selfconsistencypot}{\methodname{SC\textsubscript{PoT}}}

\newcommand{\cotmethod}{\methodname{CoT}} 
\newcommand{\potmethod}{\methodname{PoT}} 

\newcommand{\adaptiveconsistency}{\methodname{ASC}} 
\newcommand{\adaptiveconsistencycotpot}{\methodname{A\textsubscript{ASC-CP}}} 

\newcommand{\firstsecondcot}{\methodname{A\textsubscript{FS-C}}} 

\newcommand{\firstsecondpot}{\methodname{A\textsubscript{FS-P}}}

\newcommand{\cpfullmaj}{\methodname{CP\textsubscript{Maj}}}
\newcommand{\cpfullmax}{\methodname{CP\textsubscript{Max}}}
\newcommand{\cpfullagr}{\methodname{CP\textsubscript{Agr}}}

\newcommand{\cpearlybayesaa}{\methodname{CP\textsubscript{DAA}}}
\newcommand{\cpearlybayesfa}{\methodname{CP\textsubscript{DFA}}}
\newcommand{\cpearlyaa}{\methodname{CP\textsubscript{AA}}}
\newcommand{\cpearlyfa}{\methodname{CP\textsubscript{FA}}}
\newcommand{\cpearlyff}{\methodname{CP\textsubscript{FF}}}

\newcommand{\esc}{\methodname{ESC}}

In this section we describe our cross-modal ensembling approach  that combines CoT and PoT reasoning modalities  to improve both the accuracy  and efficiency of self-consistency. We explore such hybrid sampling strategies in two main directions: 1) \emph{Full sampling} approaches that use the entire available decoding budget to maximize ensemble accuracy, and  2) \emph{Early stopping} strategies that aim to minimize cost by dynamically terminating sampling once sufficient agreement between modalities is observed.

\subsection{Full sampling CoT-PoT strategies}
\label{subsec:fullsampling}

While standard self-consistency samples multiple CoT solutions and performs majority voting, in our hybrid approach we sample an equal number of CoT and PoT solutions for the full budget that is available. However, there can be multiple ways of aggregating the results across the different modalities to obtain the final answer. We consider various approaches to benefit from the complementary strengths of the two reasoning modalities.

Let \( \anscot  \) and \( \anspot  \) be the sequence of answers generated by CoT and PoT sampling for the same reasoning task, respectively. For any specific final answer \( y \), let $\freqcot{y}$ and $\freqpot{y}$ be the number of instances that reach the final answer $y$ in $\anscot$ and $\anspot$, respectively. We consider the following combination strategies:

\textbf{\cpfullmaj (Majority voting).} This approach performs a majority vote over all sampled answers from CoT and PoT combined. It leverages the overall diversity in the samples introduced by both modalities and simply takes the most frequently occurring answer. The result is obtained as: 
 \[
  y^* = \operatorname*{argmax}_y \left( \allfreq{y} \right)
  \]
\textbf{\cpfullmax (Maximum confidence modality).} In this approach we select the most frequent answer from either the CoT or PoT modality, whichever has higher frequency. This effectively lets the more confident modality dominate for each question. Formally, it is defined as:
\[
  y^* = \operatorname*{argmax}_y \left( \max\left( \freqcot{y}, \freqpot{y} \right) \right)
  \]
\textbf{\cpfullagr  (Modality agreement prioritization).} This approach first prioritizes answers that appear in both CoT and PoT, and then performs majority voting. Thus it gives the highest confidence to answers that see agreement between the two modalities. Formally, it is defined with a lexical ordering of two components: the indicator function for agreement and the total frequency for majority tie-breaking: \[
  y^*
  = \operatorname*{argmax}_{y }
  \Bigl(
    \mathbf 1\{y \in \anscot \cap \anspot\},
    \; \allfreq{y}
  \Bigr)_{\text{lex}}
  \]




\subsection{Early stopping CoT-PoT strategies}
\label{subsec:earlystopping}

While the full sampling strategies explore how accuracy may be maximized using all the available sampling budget, another important goal of this work is to leverage CoT-PoT agreement to minimize the number of samples and improve the efficiency of self-consistency techniques. Early stopping introduces a higher-stakes decision: we must decide whether to terminate sampling based on partial evidence without having seen the complete sample set. Hence, the main question is how we can measure such confidence based on agreement between the two modalities to determine when to stop. We investigate this question with a general Bayesian formulation based on the core agreement events of interest. We then describe concrete strategies that instantiate this general model, considering both \textit{data-driven} strategies that infer seed probabilities from data, and simplified \textit{data-independent} strategies based on edge-case parameterizations of the general model.

\subsubsection{Bayesian agreement model}

We have a uniform sampling scheme that alternates between generating one CoT and one PoT answer until the total sampling budget is reached. During this iterative sampling, we aim to halt early whenever there is sufficiently strong confidence based on agreement between the two reasoning modalities. We model this process with parallel Bayesian hypothesis tests that continually update as sampling proceeds. Each test is anchored to a distinct answer generated as sampling progresses. Without loss of generality, let $y$ be a unique anchor answer generated by PoT at some iteration (the CoT case is treated symmetrically). This initiates a new hypothesis test that tracks the number of CoT agreements with $y$. For this or any subsequent iteration, let $t$  be the total number of CoT answers generated so far (whether before or after $y$), and let $k\le t$ be the number of CoT answers  that match $y$. Thus, formulated as a Bernoulli process with $t$ total trials and $k$ is the number of successes,  each trial $i$ is defined as $A_{i} = 1$ if and only if $\anscot[i] = y$.

The hypothesis test for $y$ is based on the event $C$ that the answer $y$ is \emph{safe}: either $y$ is correct \emph{or}  the answer obtained from the full sampling will also be wrong. We define this as our target event of interest since we are aiming to infer high confidence for when to stop sampling, rather than strictly when the answer is correct (for which we may very rarely have very high confidence). Based on these events, the three base probabilities of interest are: 
\[
\begin{array}{l l l l}
c &\!=\!& P(C) & \text{\small $y$ is safe} \\
a_1 &\!=\!& P(A_{i}=1) & \text{\small any CoT answer equals $y$} \\
a_2 &\!=\!& P(C \mid A_{i}=1) & \text{\small $y$ is safe given agreement}
\end{array}
\]

From these we derive the likelihood of observing an agreement at trial $i$ under both hypotheses of $y$ being safe ($C$) or not ($\lnot C$):
\begin{align*}
q_1 &= P(A_{i} = 1 \mid C) = \frac{a_1 a_2}{c} \\
q_0 &= P(A_{i} = 1 \mid \lnot C) = \frac{a_1(1 - a_2)}{1 - c}
\end{align*}

Assuming independence between trials, the likelihood of observing $k$ successes in $t$ Bernoulli trials under each hypothesis is:
\begin{align*}
P(k, t \mid C) &= \binom{t}{k} q_1^k (1 - q_1)^{t - k} \\
P(k, t \mid \lnot C) &= \binom{t}{k} q_0^k (1 - q_0)^{t - k}
\end{align*}


Finally, Bayes' rule provides the posterior probability that $y$ is safe after $k$ agreements in $t$ trials: 


\begin{align*}
P(C \mid k, t) &=&\!\!\!\!\! \frac{P(C) P(k, t \mid \! C)}{P(C) P(k, t \mid \! C) + P(\lnot C) P(k, t \mid\! \lnot C)} \\
&=&\!\!\!\!\!\frac{c q_1^{k} (1 - q_1)^{t - k}}{c q_1^{k} (1 - q_1)^{t - k} + (1 - c) q_0^{k} (1 - q_0)^{t - k}}
\end{align*}

We determine an early stop as soon this posterior probability surpasses a desired confidence threshold $P(C \mid k, t) \geq \rho$. We explore two kinds of instantiation strategies of the general model above: data-driven variants that estimate model parameters from held-out agreement statistics, and heuristic variants that set extreme parameter values that reduce to simple stopping rules. In all cases the underlying sampling process samples one CoT and one PoT answer alternately, where the first CoT and PoT are generated at temperature 0 (as the LLM's most confident guess for each modality) and the rest at temperature 0.7 (for higher diversity ~\citep{wang2022self}). 

\subsubsection{Data-driven specializations} 
\label{subsubsec:datadrivenmethods}

Our general Bayesian model is parameterized by the three core probabilities $c$, $a_1$ and $a_2$. These can be inferred statistically from data by  performing full sampling and recording the rates of safety and agreement. For our data-driven strategies we infer these probabilities from the unused training sets of the benchmarks we use in our evaluation. With these inferred probabilities, we consider two specialized strategies based on which anchors are chosen to test agreements. 

\begin{description}[leftmargin=0pt, labelsep=1em]
\item[\cpearlybayesaa] This is the \emph{any-to-any} method where a new tracker is instantiated for every unique answer generated during sampling. Each tracker independently accumulates cross-modal matches, and we stop when any tracker's posterior probability exceeds the threshold.

\item[\cpearlybayesfa] This is a \emph{first-to-any} approach where we create trackers only for the two initial, temperature-0 CoT and PoT answers ($\anscot[0]$ and $\anspot[0]$). We stop as soon as either initial answer establishes agreement from the other modality. This a more conservative strategy that only anchors on the highest confidence answers from the LLM. 
\end{description}

\subsubsection{Data-independent specializations} 
\label{subsubsec:dataindependentmethods}

Empirically, a key observation in this work is that the probability of safety given agreement is generally extremely high in practice, that is, $a_2 \approx 1$. If we consider the extreme case where $a_2 = 1$, this amounts to a strategy where we stop as soon as one cross-modal agreement is seen (the posterior is always $1$ when $k = 1$ for any $t$). Based on this notion, we consider the following data-independent strategies for early-stopping.

\begin{description}[leftmargin=0pt, labelsep=1em]
\item[\cpearlyaa] This is the any-to-any approach, where we stop as soon as there is agreement between any PoT and CoT answer. This is equivalent to \cpearlybayesaa when $a_2 = 1$. 

\item [\cpearlyfa] This is the first-to-any approach, where we stop as soon as there is any cross-modal agreement with the first PoT or CoT answer. This is equivalent to \cpearlybayesfa when $a_2 = 1$.

\item[\cpearlyff] This is the most conservative first-to-first strategy, where we only test agreement between the initial temperature-0 PoT and CoT answers (thus assuming $a_2 = 1$ and $t = 1$ with only one trial). 

\end{description}

\subsubsection{Incorporating adaptive consistency}

Although cross-modal agreement is a very strong signal when it happens, it is also possible that such agreement is not observed even though one answer becomes overwhelmingly dominant for one of the modalities as sampling progresses (especially if the problem is particularly suited to a specific modality). To capture this complementary evidence, we incorporate the adaptive consistency approach~\citep{adaptiveconsistency} in all of our early-stopping strategies. This is done by including another parallel hypothesis test that implements the Beta-stopping rule of~\cite{adaptiveconsistency} in our alternating CoT-PoT sampling process. Let $v_1$ and $v_2$ be the current vote counts of the most frequent and the second-most frequent answers, respectively, aggregated over \emph{all} CoT and PoT samples observed so far. Assuming a uniform $\operatorname{Beta}(1,1)$ prior on the true share~$\theta$ of the leading answer, the posterior is modelled as  $ \operatorname{Beta}(v_1+1,\;v_2+1)$, and the probability that the leader will remain the majority after unlimited additional sampling is simply the tail mass of this posterior Beta distribution above 0.5. We perform this Beta majority test in parallel with every cross-stream tracker and terminate as soon as any of these tests exceed the confidence threshold.  

%% file: evaluation.tex
\section{Evaluation}
\label{sec:evaluation}

\newcommand{\datasetname}[1]{\textsc{#1}\xspace}

\newcommand{\gsm}{\datasetname{GSM8k}}
\newcommand{\mathdataset}{\datasetname{MATH}}
\newcommand{\svamp}{\datasetname{SVAMP}}
\newcommand{\finqa}{\datasetname{FinQA}}
\newcommand{\tabwmp}{\datasetname{TabMWP}}

\newcommand{\modelname}[1]{\textsc{#1}\xspace}

\newcommand{\gptthreepointfive}{\datasetname{GPT-3.5}}
\newcommand{\gptfouromni}{\datasetname{GPT-4o}}
\newcommand{\mistral}{\datasetname{Mistral-Lrg}}
\newcommand{\qwen}{\datasetname{Qwen3-Coder}}
\newcommand{\deepseek}{\datasetname{DeepSeek-R1}}

\newcommand{\llamabase}{\datasetname{Llama3B}}
\newcommand{\llamatrained}{\datasetname{Llama3B-CP}}

\begin{table}[t]
\centering
\fontsize{9}{11}\selectfont
\setlength{\tabcolsep}{4pt}
\begin{tabular}{lccc}
\toprule
\textbf{Model} & \(c\) & \(a_{1}\) & \(a_{2}\) \\
\midrule
\gptthreepointfive  & 0.840 & 0.527 & 0.996 \\
\gptfouromni        & 0.862 & 0.667 & 0.999 \\
\mistral            & 0.861 & 0.682 & 0.996 \\
\qwen             & 0.938 & 0.844 & 0.998 \\
\deepseek         & 0.918 & 0.783 & 0.991 \\
\bottomrule
\end{tabular}
\caption{Inferred parameter probabilities}
\label{tab:parameter-probabilities}
\end{table}

In this section we present an evaluation of both the accuracy and efficiency of our full-sampling and early-stopping CoT-PoT ensembling methods.

 \textbf{\emph{Datasets.}} We use five benchmarks covering a range of different kinds of reasoning tasks:
{\bf {GSM8K}} (\citealp{cobbe2021training}) consists of elementary to middle school level  word problems; {\bf {MATH}} (\citealp{hendrycksmath2021}) consists of challenging high-school level math competition problems covering advanced topics including algebra, calculus and geometry; {\bf {FinQA}} (\citealp{chen2021finqa}) contains  problems from real-world financial contexts, requiring integrated reasoning over textual and structured data; {\bf {SVAMP}} (\citealp{patel2021nlp}) contains arithmetic word problems designed to identify common numerical reasoning pitfalls in NLP models; and {\bf {TabMWP}} (\citealp{lu2022dynamic}) contains semi-structured problems involving reasoning with text and tabular data. For our evaluation we use 500 cases from the test splits of each dataset (to cap costs of our large sampling experiments).

\begin{table*}[t]
\centering
\fontsize{9}{11}\selectfont
\setlength{\tabcolsep}{5pt}
\begin{tabular}{l c c ccc c c}
\toprule
\;Dataset & \selfconsistency & \selfconsistencypot & \cpfullmaj & \cpfullmax & \cpfullagr & \cotmethod & \potmethod \\
\cmidrule(lr){1-1} \cmidrule(lr){2-8}
\;\finqa   & 70.4 & 71.7 & 72.1 & \textbf{72.2} & 72.0 & 68.0 & 70.0 \\
\;\tabwmp  & 88.2 & 86.3 & \textbf{88.4} & \textbf{88.4} & \textbf{88.4} & 87.1 & 81.6 \\
\;\svamp   & 94.8 & 94.3 & \textbf{95.6} & \textbf{95.6} & 95.5 & 93.0 & 93.2 \\
\;\gsm     & 95.7 & 93.9 & \textbf{96.1} & 96.0 & 96.0 & 93.2 & 91.8 \\
\;\mathdataset & 74.2 & 68.2 & 76.0 & \textbf{76.3} & 75.8 & 66.0 & 55.8 \\
\\[-0.7em]
\toprule
\;Model & \multicolumn{6}{c}{} \\
\cmidrule(lr){1-1}
\;\gptthreepointfive & 75.6 & 70.6 & \textbf{77.3} & 77.2 & 77.0 & 68.4 & 66.1 \\
\;\gptfouromni       & 86.3 & 84.8 & \textbf{87.2} & \textbf{87.2} & 87.0 & 83.9 & 79.5 \\
\;\mistral           & 85.0 & 84.0 & 86.1 & \textbf{86.3} & 86.1 & 82.7 & 80.0 \\
\;\qwen              & 87.7 & 87.0 & 88.3 & \textbf{88.4} & \textbf{88.4} & 85.6 & 83.6 \\
\;\deepseek          & 88.6 & 87.9 & \textbf{89.3} & \textbf{89.3} & 89.2 & 86.6 & 83.3 \\
\\[-0.7em]
\toprule
\;Average            & 84.6 & 82.9 & 85.6 & \textbf{85.7} & 85.5 & 81.4 & 78.5 \\
\bottomrule
\end{tabular}
\caption{Accuracy (\%) of full sampling methods across datasets and models.}
\label{tab:full_sampling_results}
\end{table*}

\textbf {\emph{Models.}} We evaluate our methods over five different mainstream large language models: \textbf{GPT‑3.5‑Turbo} (\citealp{openai2022_gpt35turbo}) and the more powerful \textbf{GPT‑4-Omni} (\citealp{openai2024_gpt4o}) models from OpenAI;
\textbf{Mistral‑Large} (\citealp{mistral2024_large}), which is a 123B parameter competitive reasoning model  from Mistral; \textbf{Qwen3-Coder}, which is 30B parameter open-source model with state-of-the-art coding capabilities \citep{qwen3}: and \textbf{DeepSeek-R1}, which is an advanced open-source reasoning model from DeepSeek with 671B parameters \cite{guo2025deepseek}.

\textbf{\emph{Sampling parameters.}} In our sampling process we generate a maximum of 40 samples, as in prior work \citep{wang2022self,adaptiveconsistency}. For CoT-only or PoT-only baseline methods, all of these are either CoT samples or PoT samples. For our CoT-PoT methods, we sample one CoT and one PoT response alternately until the maximum budget is reached. For all methods, including baselines, the first samples for each modality are taken at temperature 0 and the rest at 0.7. Our CoT and PoT prompts are shown in  Appendix \ref{sec:prompts}. To ensure uniformity, we used the same few-shot example questions for both CoT and PoT prompts, with PoT prompts generally requiring fewer tokens. Finally, we use  a confidence threshold of $\rho =0.975$ for our early-stopping CoT-PoT hypothesis tests.

\textbf{\emph{Seed probabilities inference.}} As discussed in Section \ref{subsubsec:datadrivenmethods}, for our data-driven early stopping strategies  we used held‑out data to infer the three parameter probabilities of our Bayesian model: $c$ (safety), $a_1$ (agreement) and $a_2$ (safety given agreement). We randomly sampled \(100\) problems from the training split of each of our datasets, performed full CoT-PoT ensembling, and  computed maximum‑likelihood estimates for each of the agreement events. These are  used as the priors in all our data-driven strategies. Table \ref{tab:parameter-probabilities} shows the inferred average probabilities for each language model. In particular, we note the consistently high values for  $a_2 \approx 1$, which  illustrates the very strong empirical correlation between cross‑modal agreement and answer safety (either the answer is correct or full sampling will also not yield the correct answer). 


\subsection{Full sampling results}

The results of our full sampling CoT-PoT strategies are shown in Table \ref{tab:full_sampling_results}. This shows the accuracy of each of our full sampling methods defined in Section \ref{subsec:fullsampling}. We compare these against the standard self-consistency approach ~\citep{wang2022self}, with the baseline methods \selfconsistency and \selfconsistencypot that perform a majority vote on CoT-only and PoT-only samples respectively. We also include for comparison the direct \cotmethod and \potmethod baselines that represent the single temperature-0 sample from each modality. The table shows the results for each dataset (averaged across models), for each model (averaged across all datasets), as well as overall average.

The main result is that {\ul{\emph{CoT-PoT ensembling has higher accuracy than both CoT-only and PoT-only self-consistency}}}. All the CoT-PoT methods perform better than \selfconsistency and \selfconsistencypot, with an overall average accuracy increase of 1.1\%. Moreover, while CoT-only and PoT-only methods outperform each other on specific datasets, the CoT-PoT methods  consistently perform better than both baselines on each of the five datasets  and five language models. Although we do not observe a significant difference between the three CoT-PoT aggregation strategies, maximization within modalities (\cpfullmax) overall performs slightly better than general majority voting (\cpfullmaj) and inter-modality agreement (\cpfullagr). This indicates that confidence within modalities provides some additional benefit over general consensus  between them.


\subsection{Early-stopping sampling results}

\begin{table*}[t]
\centering
\fontsize{9}{11}\selectfont
\setlength{\tabcolsep}{5pt}
\begin{tabular}{l*{7}{c}|*{7}{c}}
\toprule
\multicolumn{1}{c}{} & \multicolumn{7}{c}{\textbf{Accuracy}} & \multicolumn{7}{c}{\textbf{Number of samples}} \\
\cmidrule(lr){2-8} \cmidrule(lr){9-15}
\textbf{Dataset} &
\adaptiveconsistency & \esc & \cpearlyaa & \cpearlyfa & \cpearlyff & \cpearlybayesaa & \cpearlybayesfa \quad&
\adaptiveconsistency & \esc & \cpearlyaa & \cpearlyfa & \cpearlyff & \cpearlybayesaa & \cpearlybayesfa \\
\cmidrule(lr){1-1}
\cmidrule(lr){2-8}
\cmidrule(lr){9-15}
\gsm
& 95.7 & 95.8 & 95.7 & 95.8 & \textbf{96.2} & 95.9 & 95.8 \quad&
5.3 & 6.6 & \textbf{2.6} & 2.8 & 3.7 & 2.7 & 3.2 \\
\mathdataset
& 74.2 & 74.1 & 75.8 & 75.6 & 76.0 & \textbf{76.1} & 75.8 \quad&
12.6 & 16.4 & \textbf{8.1} & 9.8 & 12.6 & 8.3 & 10.4 \\
\svamp
& 94.7 & 94.8 & 95.0 & 95.3 & \textbf{95.6} & 95.0 & 95.5 \quad&
5.3 & 6.6 & \textbf{2.5} & 2.6 & 3.3 & \textbf{2.5} & 2.8 \\
\finqa
& 70.4 & 70.4 & 71.9 & 72.0 & 72.0 & 71.9 & \textbf{72.1} \quad&
6.9 & 9.1 & \textbf{3.9} & 4.5 & 6.1 & \textbf{3.9} & 4.8 \\
\tabwmp
& 88.3 & 88.2 & 88.3 & \textbf{88.4} & \textbf{88.4} & 88.3 & 88.3 \quad&
5.8 & 6.8 & \textbf{4.3} & 4.5 & 5.5 & 4.4 & 4.7 \\
\\[-0.7em]
\toprule
\textbf{Model} & \multicolumn{13}{c}{}  \\
\cmidrule(lr){1-1}
\gptthreepointfive
& 75.6 & 75.6 & 76.7 & 76.8 & \textbf{77.4} & 77.0 & 76.9 \quad&
11.7 & 16.5 & \textbf{6.0} & 7.4 & 10.9 & 6.2 & 8.4 \\
\gptfouromni
& 86.3 & 86.2 & 86.8 & 86.8 & \textbf{87.2} & 86.8 & 86.8 \quad&
6.3 & 8.6 & \textbf{3.9} & 4.3 & 5.4 & \textbf{3.9} & 4.4 \\
\mistral
& 85.0 & 85.0 & 85.9 & 85.9 & 86.0 & 86.0 & \textbf{86.2} \quad&
6.7 & 9.0 & \textbf{3.9} & 4.4 & 5.7 & 4.0 & 4.8 \\
\qwen
& 87.7 & 87.8 & 88.4 & \textbf{88.5} & 88.3 & 88.4 & 88.4 \quad&
5.9 & 4.4 & \textbf{3.8} & 4.0 & 4.9 & \textbf{3.8} & 4.3 \\
\deepseek
& 88.6 & 88.6 & 89.1 & \textbf{89.3} & \textbf{89.3} & 89.1 & \textbf{89.3} \quad&
5.3 & 7.0 & \textbf{3.8} & \textbf{3.8} & 4.5 & \textbf{3.8} & 4.0 \\
\\[-0.7em]
\toprule
\textbf{Average}
& 84.6 & 84.6 & 85.4 & 85.4 & \textbf{85.6} & 85.5 & 85.5 \quad&
7.2 & 9.1 & \textbf{4.3} & 4.8 & 6.3 & 4.4 & 5.2 \\
$\Delta$\textbf{-\cpfullmax}
& -1.1 & -1.1 & -0.3 & -0.3 & \textbf{-0.1} & -0.2 & -0.2 \quad&
5.6x & 4.4x & \textbf{9.3x} & 8.3x & 6.3x & 9.1x & 7.7x \\
$\Delta$-\selfconsistency
& 0.0 & 0.0 & +0.8 & +0.9 & \textbf{+1.0} & +0.9 & +0.9 \quad&
5.6x & 4.4x & \textbf{9.3x} & 8.3x & 6.3x & 9.1x & 7.7x \\
\bottomrule
\end{tabular}

\caption{Accuracy and number of samples for early-stopping strategies. $\Delta$ rows show the difference in accuracy and factor of reduction in the number of samples in comparison to full-sampling methods.}
\label{tab:early_stopping_results}
\end{table*}

Table \ref{tab:early_stopping_results} shows both the accuracy and efficiency (number of samples required) for each of our early-stopping CoT-PoT methods defined in Sections \ref{subsubsec:datadrivenmethods} and \ref{subsubsec:dataindependentmethods}. We compare our methods against the two prior state-of-the-art early stopping approaches that use CoT-only sampling: \adaptiveconsistency is the adaptive consistency method based on early majorities ~\citep{adaptiveconsistency}, and \esc is the early-stopping approach based on sampling windows \citep{escapeskyhighcost}. We make the following key observations:

1.{\ul{ \emph{Every CoT-PoT method provides higher accuracy and efficiency than all prior early-stopping and full-sampling methods}}}. Our most efficient early-stopping method is \cpearlyaa. This method provides the biggest efficiency improvement among all methods, while still  having $0.8$\% higher accuracy than full sampling \selfconsistency and both early-stopping baselines. Overall, it reduces the number of samples drastically by a factor of  $9.3\times$  as compared to $5.6\times$ by the best prior early-stopping method  \adaptiveconsistency. It also consistently shows the best efficiency for each of the datasets and models. On the other hand, our most accurate early-stopping method is the conservative approach of \cpearlyff. While having a higher accuracy than all prior early-stopping and full sampling methods, and only a $0.1$\% accuracy drop compared to our most accurate full sampling method (\cpfullmax), this method still provides better efficiency than the prior early-stopping methods, reducing sampling by $6.3\times$  as compared to $5.6\times$ by \adaptiveconsistency and $4.4\times$ by \esc. 

2.{\ul{ \emph{The data-driven CoT-PoT methods perform best for certain models and domains}}}. While on average the most efficient and most accurate methods are \cpearlyaa and \cpearlyff,  for certain models and datasets the data-driven methods provide the best results. For instance, \cpearlybayesfa has the highest accuracy for the Mistral model (across all datasets) and on the FinQA dataset (finance domain questions across all models) while also providing better efficiency than \cpearlyff. This method also matches the highest accuracy for the R1 model. Similarly, \cpearlybayesaa provides the best accuracy for the MATH dataset (advanced maths topics) while providing significantly higher efficiency than \cpearlyff. Thus for certain models and domains we see that the simple choice of first or any CoT-PoT agreement is sub-optimal, and learning CoT-PoT confidence from data provides the best results. In general, the data-driven  methods are more accurate and less efficient than their data-independent counterparts. This can be expected as our most efficient (\cpearlyaa) and most accurate (\cpearlyff) methods represent the most aggressive and  conservative extremes of our CoT-PoT approach on average, while the data-driven methods generally provide trade-offs between accuracy and efficiency. The data-independent methods present simpler alternatives to leverage  CoT-PoT consistency easily without the need for  parameter inference from data.

\begin{figure}[t]
  \centering
  \includegraphics[width=0.99\columnwidth]{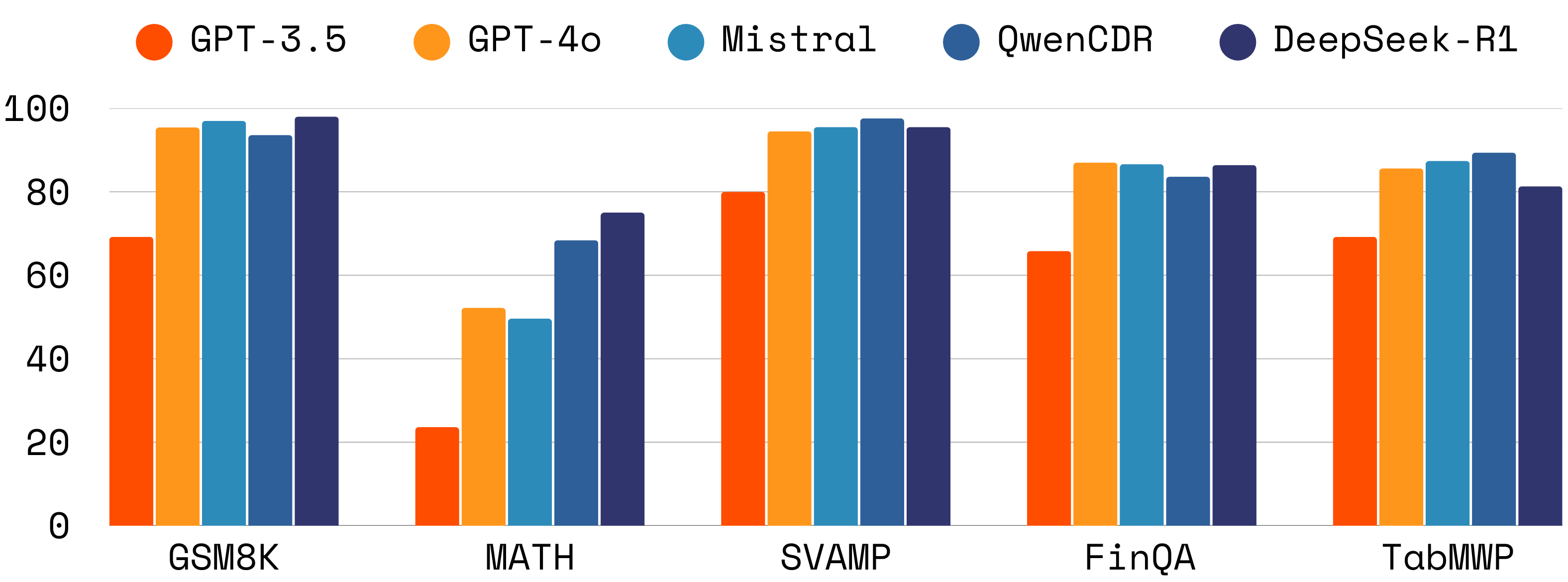} 
  \caption{Percentage of problems solved with only two samples by early-stopping CoT-PoT methods.}
  \label{fig:two-sample-rate}
\end{figure}

3.{\ul{ \emph{Early-stopping with CoT-PoT requires only two samples in the majority of cases}}}. Figure \ref{fig:two-sample-rate} shows the percentage of test cases that can be solved with just two samples (one CoT and one PoT) by our CoT-PoT methods (this percentage is the same for all the CoT-PoT early-stopping methods). Overall, on average across all models and datasets, CoT-PoT methods can terminate with just 2 samples in $78.6\%$ of cases. No prior technique can terminate with only two samples in any scenario: \adaptiveconsistency requires a minimum of 4 and \esc requires at least 5 samples in all cases. We also observe that termination from 2 samples is more common for the stronger reasoning models (e.g. R1) as opposed to weaker models (e.g. GPT-3.5) within each domain, which illustrates how cross-modal consistency is more practically beneficial for more powerful reasoning-intensive models.

4.{\ul{ \emph{CoT-PoT efficiency gains hold under token-level and latency measurement}}}. While the sample count results above already demonstrate strong efficiency gains, we also validate these findings using token-level usage as a more fine-grained measure of computational cost. Table~\ref{tab:token_efficiency} reports the average prompt+completion tokens (in thousands) per method, across all models and datasets. The overall pattern mirrors the sample count results: \cpearlyaa achieves the highest token efficiency overall, with a $9.8\times$ reduction in token usage compared to CoT-only full sampling (\selfconsistency), exceeding even the $9.3\times$ sample count reduction, since PoT prompts tend to be shorter than CoT prompts. All our CoT-PoT methods consistently outperform the early-stopping baselines \adaptiveconsistency ($5.4\times$) and \esc ($3.8\times$) in token efficiency.

We also measured the overhead that comes from executing the programs created by the PoT modality. The average execution time was $422.5$~ms per PoT program across all datasets. For the $78.6\%$ of cases where \cpearlyaa terminates after just one CoT and one PoT sample, this adds under half a second of execution latency. On average, \cpearlyaa requires $4.3$ total samples (Table~\ref{tab:early_stopping_results}), corresponding to roughly 2 PoT executions per query and approximately ${\sim}900$~ms of total execution overhead. This is negligible relative to the savings from reducing LLM forward passes from 40 to 4.3 on average. Moreover, PoT execution is also fully parallelizable.


\begin{table*}[t]
\centering
\fontsize{8}{10}\selectfont
\setlength{\tabcolsep}{4pt}
\begin{tabular}{l*{10}{c}}
\toprule
 & \cpfullmax & \selfconsistency & \selfconsistencypot & \adaptiveconsistency & \esc & \cpearlyaa & \cpearlyfa & \cpearlyff & \cpearlybayesaa & \cpearlybayesfa \\
\midrule
\textbf{Dataset} \\
\;\gsm        & 33.0 & 36.5 & 29.4 & 4.8 & 6.6 & \textbf{2.1} & 2.2 & 3.0 & 2.2 & 2.5 \\
\;\mathdataset & 54.7 & 60.0 & 49.1 & 18.3 & 26.0 & \textbf{11.0} & 13.0 & 16.3 & 11.2 & 14.0 \\
\;\svamp      & 29.3 & 33.3 & 25.2 & 4.6 & 6.2 & \textbf{1.9} & 2.0 & 2.5 & \textbf{1.9} & 2.1 \\
\;\finqa      & 247.4 & 252.9 & 241.6 & 44.2 & 63.4 & \textbf{23.6} & 25.2 & 37.8 & 23.9 & 29.5 \\
\;\tabwmp     & 27.4 & 25.8 & 29.0 & 3.6 & 4.7 & \textbf{2.9} & 3.0 & 3.6 & \textbf{2.9} & 3.1 \\
\midrule
\textbf{Model} \\
\;\gptthreepointfive & 65.6 & 69.2 & 62.0 & 19.1 & 28.8 & \textbf{9.5} & 11.2 & 17.3 & 9.7 & 12.5 \\
\;\gptfouromni       & 70.2 & 76.0 & 63.9 & 12.5 & 17.3 & \textbf{6.4} & 6.9 & 9.6 & \textbf{6.4} & 7.5 \\
\;\mistral           & 71.0 & 73.4 & 68.6 & 13.2 & 17.7 & \textbf{6.5} & 7.5 & 10.3 & 6.7 & 8.5 \\
\;\qwen              & 73.8 & 75.9 & 71.7 & 12.7 & 17.8 & \textbf{8.1} & 8.5 & 11.4 & 8.3 & 10.5 \\
\;\deepseek          & 111.2 & 114.1 & 108.3 & 18.1 & 25.4 & \textbf{10.9} & 11.4 & 14.5 & \textbf{10.9} & 12.2 \\
\midrule
\textbf{Average}     & 78.4 & 81.7 & 74.9 & 15.1 & 21.4 & \textbf{8.3} & 9.1 & 12.6 & 8.4 & 10.2 \\
$\Delta$-\cpfullmax  & 1.0x & 1.0x & 1.1x & 5.2x & 3.7x & \textbf{9.5x} & 8.6x & 6.2x & 9.3x & 7.7x \\
$\Delta$-\selfconsistency & 1.0x & 1.0x & 1.1x & 5.4x & 3.8x & \textbf{9.8x} & 9.0x & 6.5x & 9.7x & 8.0x \\
\bottomrule
\end{tabular}
\caption{Average prompt+completion tokens (thousands) per method, across datasets and models. $\Delta$ rows show token efficiency gain relative to full-sampling baselines. \cpearlyaa achieves a $9.8\times$ token reduction versus \selfconsistency, exceeding even the sample-count gain of $9.3\times$.}
\label{tab:token_efficiency}
\end{table*}

\subsection{Ablation studies}



We performed ablation experiments to further examine different aspects of our cross-modal  approach. Firstly, we evaluated the benefit of early stopping using cross-modal agreement versus only using  adaptive consistency  over a mix of CoT and PoT samples. We test this with the \adaptiveconsistencycotpot ablation, which is adaptive consistency applied over our hybrid sampling scheme of alternating CoT and PoT samples (rather than CoT-only samples as in  \adaptiveconsistency). Table \ref{tab:ablations} shows the ablation results: while \adaptiveconsistencycotpot reaches similar accuracy as our best CoT-PoT early stopping methods, it requires significantly more samples: 8.2 samples in comparison to 4.3 for \cpearlyaa and 6.3 for \cpearlyff. This shows how cross-modal agreement provides further efficiency gains over just performing adaptive consistency on a mixture of CoT and PoT samples. 

\begin{table}[t]
\centering
\fontsize{9}{11}\selectfont
\setlength{\tabcolsep}{3pt}
\begin{tabular}{l*{6}{c}}
\toprule
& \cpfullmax & \cpearlyaa & \cpearlyff & \adaptiveconsistencycotpot & \firstsecondcot & \firstsecondpot \\
\midrule
{\bf Accuracy}  & 85.68 & 85.36 & 85.64 & 85.65 & 84.03 & 80.77 \\
{\bf \#Samples} & 40.0 &  4.3 & 6.3 & 8.2 & 6.4 &  6.9 \\
\bottomrule
\end{tabular}
\caption{Ablations compared to CoT-PoT methods.}
\label{tab:ablations}
\end{table}

We also examined how cross-modal agreement compares to within-modality agreement. We define the ablation  \firstsecondcot as standard CoT-only self-consistency but where an answer is returned early if there is agreement between the first and second CoT samples. \firstsecondpot is defined similarly but with PoT-only samples. These methods are in contrast to \cpearlyff which stops on agreement between the first CoT and PoT samples. The results in Table \ref{tab:ablations} show that both these methods have a significantly lower accuracy than the CoT-PoT early stopping methods while also being less sample-efficient. They also show lower accuracy than their respective full sampling counterparts \selfconsistency and \selfconsistencypot (as shown in Table \ref{tab:full_sampling_results}). This indicates the robustness of cross-modal  agreement as an early stopping signal in comparison to within-modality agreement for either modality.


\begin{figure}[t]
  \centering
  \includegraphics[width=0.99\columnwidth]{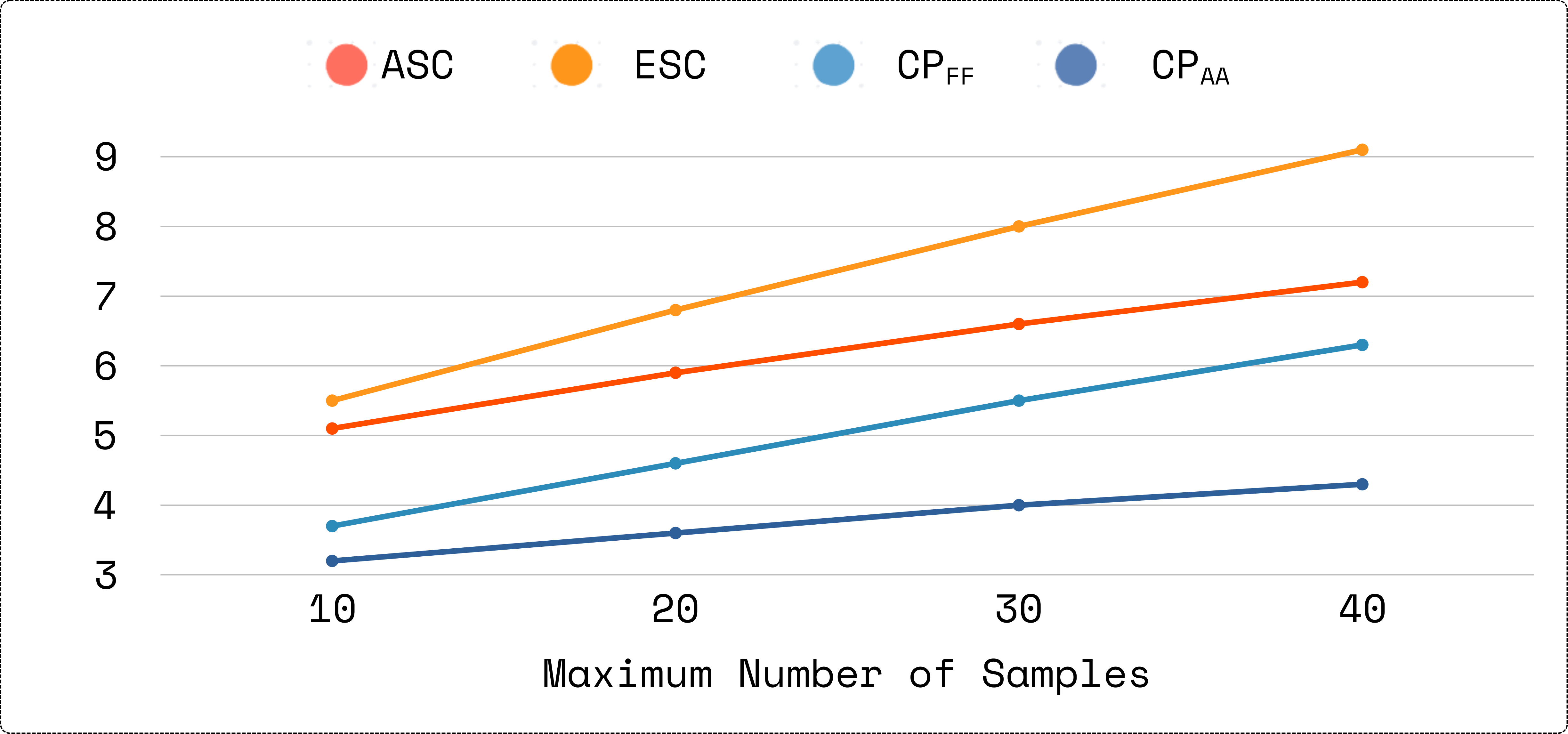}
  \caption{ Number of samples used (Y-axis) vs.\ maximum sampling budget across different early-stopping methods (lower values indicate higher efficiency).}
  \label{fig:efficiency-samplingincrease}
\end{figure}

 We also investigated how the efficiency of early-stopping methods changes with   sampling budget. Figure \ref{fig:efficiency-samplingincrease} shows the number of samples used  as the maximum sampling budget increases from 10 to 40. While all methods utilize more samples as the budget increases, we observe a greater gap between the baselines and our most efficient method \cpearlyaa as budget increases. This shows how  the efficiency gains with CoT-PoT ensembling increase with  higher sampling budgets.

\paragraph{Failure Cases}
While CoT–PoT agreement is a strong indicator of correctness, we discuss here the rare cases where the two modalities may agree on an incorrect answer. We show examples of such cases across different models and datasets in Appendix \ref{sec:failure_cases}. In general,  we observe that in such cases both  modalities adopt the same \emph{conceptual misinterpretation} of the problem, while otherwise exhibiting internally consistent reasoning and computation. For example, in many instances both modalities misinterpret the semantic context of the question (e.g. considering the previous day’s quantity rather than the current one, treating a break-even year as the first profitable year) leading to off-by-one or boundary-condition errors. In other cases, both CoT and PoT rely on an incorrect but plausible abstraction (e.g. assuming a one-time salary increase rather than a recurring one, misaligning columns in a table, or treating digit sums as sufficient for modular arithmetic). In these cases, the PoT  encodes the same flawed conceptual model used by the CoT, yielding agreement with a shared but incorrect problem understanding. We also note that when CoT–PoT agreement fails in this manner, full self-consistency sampling also  most often fails, as we see in these sample cases as well as the extremely high  probability of safety given agreement that we have observed empirically ($a_2$ in Table \ref{tab:parameter-probabilities}). This indicates that self-consistency sampling mostly helps to reduce stochastic, arithmetic, or execution-level errors, as opposed to more conceptual or interpretational biases.



%% file: selftraining.tex
\subsection{Case Study: Bootstrapping PoT from CoT}
\label{sec:selftraining}

\begin{table*}[t]
\centering
\small
\setlength{\tabcolsep}{4pt}
\begin{tabular}{l*{8}{c}|*{4}{c}}
\toprule
\multicolumn{1}{c}{} & \multicolumn{8}{c}{\textbf{Accuracy}} & \multicolumn{4}{c}{\textbf{Number of samples}} \\
\cmidrule(lr){2-9} \cmidrule(lr){10-13}
\textbf{Model} &
\cotmethod & \potmethod & \selfconsistency & \cpfullmaj & \adaptiveconsistency & \cpearlyaa & \cpearlyff & \cpearlybayesfa \;&
\; \adaptiveconsistency & \cpearlyaa & \cpearlyff & \cpearlybayesfa \\
\cmidrule(lr){1-1}
\cmidrule(lr){2-9} \cmidrule(lr){10-13}
\llamabase     & 35.6 & 19.2 & 46.6 & 48.4 & 46.6 & 47.6 & 48.4 & 47.4 \quad &
\; 13.5 & 10.4 & 14.7 & 12.7 \\
\llamatrained  & 36.4 & 32.4 & 51.4 & 52.8 & 51.4 & 49.8 & 52.6 & 51.6 \quad &
\; 14.2 & 8.4 & 13.8 & 11.8 \\
\bottomrule
\end{tabular}
\vspace{-5pt}
\caption{Accuracy and efficiency before and after SFT with PoT self-induction on Llama 3B model.}
\vspace{-5pt}
\label{tab:llama_training}
\end{table*}

While large commercial-scale models generally exhibit CoT and PoT modalities well as seen in our main results, smaller models may often lack strong PoT capabilities. Moreover, PoT training data may not be readily available, and prior work often relies on expensive teacher models like GPT-4 to generate PoT rationales~\citep{mammoth, tora}. To investigate this limitation, we conducted a case study on a small 3B-parameter Llama 3.2 model on the most challenging \mathdataset benchmark. PoT  was indeed  significantly weaker than CoT in this model (19.2\% vs. 35.6\%), though CoT-PoT still outperformed standard SC. To improve PoT performance, we explored a \emph{bootstrapping} approach where PoT reasoning was self-induced from CoT: using questions and the CoT data from 4000 problems from the train set, we prompted the base model to generate PoT rationales (prompt in Appendix \ref{sec:prompts}). Only outputs matching the ground truth were retained, and we iteratively fine-tuned on these to generate further PoT rationales, yielding 2790 in total. This is similar to self-training methods \citep{zelikman2022starbootstrappingreasoningreasoning}, except that we inferred one reasoning modality from another. Finally, we used all of the self-generated PoT data to fine-tune  a lightweight LoRA adapter for PoT on the base model and trained a similar CoT adapter using the original CoT data. This enabled a simple switch between modalities: during ensembling, the relevant adapter was activated when sampling from each modality. 

Table ~\ref{tab:llama_training} presents results for both the base model and the CoT-PoT enhanced model on the MATH test set. Firstly, even on the weak base  model, the CoT-PoT methods outperformed prior methods in both accuracy and efficiency. With the LoRA adapters enabled, there was significant PoT improvement, which led to larger accuracy improvements across all CoT-PoT ensembling  strategies over CoT-only ensembling (even though there was less bootstrapped data for PoT training than for CoT). However, we notice that in this setting the data-independent \cpearlyaa method had lower accuracy than  \selfconsistency (49.8 vs. 51.4), but its data-driven counterpart \cpearlybayesfa is the most efficient method that has higher accuracy than self-consistency baseline (51.6). This shows the robustness gain from data-driven inference in this weaker model setting where the heuristic of the first cross-modal agreement is less reliable. Overall, this study illustrates an interesting research direction where one reasoning modality  can be effectively improved by another to provide overall cross-modal ensembling benefits.

%% file: relatedwork.tex
\section{Related Work}
\label{sec:relatedwork}

\textbf{Test Time Scaling for LLMs.} 
Test-time scaling strategies fall into two main categories: sequential refinement and parallel sampling. Sequential methods, such as long chains of thought~\cite{openai_learning_to_reason_2024, guo2025deepseek} and self-correction~\citep{huang2022large, madaan2024self, lee2025evolving}, guide models through multi-step reasoning and revision. While widely adopted in recent models, much of their development focuses on training-time integration. In contrast, parallel approaches like Best-of-$N$ improve solution coverage by generating multiple responses~\citep{chollet2019measure, irvine2023rewarding, brown2024large}, though selecting the correct solution remains challenging~\citep{brown2024large, hassid2024larger, christiano2017deep, wang2024math}.  In this context, self-consistency (SC) has emerged as an effective tool for test-time inference \citep{wang2022self}.  In this work, we explore improving both the accuracy and efficiency of SC via cross-modal reasoning.

\textbf{Improving efficiency of self-consistency.} While SC improves accuracy, it comes with high inference-time cost, and various approaches have been proposed to improve its efficiency. 
Adaptive consistency uses early stopping when a confident majority emerges during sampling~\citep{adaptiveconsistency}.   A related approach~\citep{escapeskyhighcost} checks for early majority within  small windows, though this always requires a fixed minimal sample size.  \citep{difficultyadaptiveconsistency} propose allocating sampling budgets based on problem difficulty, but this only works over large problem batches to infer relative difficulty. All of these methods operate within the single CoT modality and depend on majority or difficulty estimation. In contrast, we propose a cross-modal ensembling approach that leverages agreement between CoT and PoT reasoning as a strong early stopping signal. We have shown how  our approach provides both higher accuracy and drastic efficiency gains over prior approaches, including inference from only two samples, which no prior method can provide.

\textbf{Combining Chain-of-Thought and Program-of-Thought.}
Recent work has investigated combinations of CoT and PoT reasoning in various ways. \citet{mammoth} show the value of fine-tuning models on both CoT and PoT data and using the two modalities for different kinds of problems. Other approaches devise specialized prompting or selection mechanisms that incorporate CoT and PoT in different ways, such as question generalization \citep{mathprompter} or assigning different reasoning modalities to different problem types \citep{xofthought}. Most directly related to our work, \citet{automaticmodelselection23} propose an automatic model selection method that uses additional LLM calls with a specialized selection prompt to choose between CoT and PoT solutions. However, this approach is not a self-consistency or early-stopping method: it requires at least 10 LLM calls per question (and up to 15) with no mechanism for early termination, and their most efficient reported variant still underperforms CoT SC with 40 samples in accuracy. This is less sample-efficient than even existing early-stopping baselines such as ASC and ESC. LLM cascades \citep{llmcascades} use large numbers of CoT and PoT inferences as a gating signal to decide whether a task should be solved by a weaker model or escalated to a stronger one. This approach relies on a fixed and large set of samples (around 20) to make the gating decision, and does not demonstrate that ensembling across modalities yields higher accuracy or reduces the number of samples required for reasoning.  Hence, the primary novelty of all of these methods lies in \emph{how} they specialize CoT and PoT for their particular goals, but none incorporates them within self-consistency to provide early stopping. In contrast, our work's primary contribution is \emph{cross-modal self-consistency with reliable early stopping}: we show that agreement between CoT and PoT provides a strong enough signal that the majority of problems (78.6\%) can be resolved with only two samples while still providing higher accuracy than full sampling, which no prior work has shown.

%% file: conclusion.tex
\section{Conclusion}
\label{sec:conclusion}

We have presented a cross-modal ensembling approach that combines Chain-of-Thought and Program-of-Thought reasoning to improve both the accuracy and efficiency of self-consistency in LLMs. The primary novelty of our work is in leveraging cross-modal agreement as a reliable early-stopping signal: because CoT and PoT exhibit different error modes with low correlation, agreement between them provides a much stronger correctness signal than within-modality agreement. Our experiments across diverse benchmarks and models show that our CoT-PoT early-stopping approach consistently outperforms all prior self-consistency methods. In particular, it provides a drastic improvement in efficiency, requiring only two samples in most cases, which has not been possible with any prior technique.



%% file: limitations.tex


\section{Limitations}
\label{sec:limitations}

While our CoT-PoT ensembling approach demonstrates significant improvements in both accuracy and efficiency over existing self-consistency methods, we note the following potential limitations:

\begin{itemize}
    \item \textbf{Applicability to programmatic tasks:} Our method is most effective for tasks where programmatic or symbolic reasoning is applicable, such as mathematical or algorithmic problems. This aligns with the large class of reasoning tasks targeted by any program-based or tool-augmented reasoning methods in the literature, e.g. \cite{tora,mammoth,potpaper,palpaper}. For tasks that are less  naturally expressible as programs (e.g. long-form QA, subjective reasoning), the PoT channel may contribute less signal but the framework will still function.  For example, preliminary experiments on broader applicability of our approach to the multi-hop question answering benchmark  HotpotQA~\citep{yang-etal-2018-hotpotqa} show improved performance with our CoT-PoT method versus CoT-only SC. We observed in such cases the LLM tends to embed CoT-like reasoning in PoT code comments even for text-heavy questions, and PoT samples can still provide partial structure (e.g., arithmetic steps, helper variables). This suggests that the cross-modal signal may have benefits beyond purely numerical domains. Future work can also explore adapting our cross-modal framework to detect when tasks may benefit from different kinds of reasoning modalities and adjust the sampling distribution accordingly.  
    
    \item \textbf{Early-stopping of data-independent methods:} Our most efficient and data-independent early-stopping method relies on any CoT–PoT agreement as a strong signal of correctness.  While we show this to be effective across all the mainstream  models and benchmarks, we have also shown cases where data-dependent methods provide more robustness, such as in the weaker Llama 3B model setting.  Thus in general, the any-any agreement approach may not be the most accurate for all kinds of tasks or application domains. In such cases, our more conservative and data-dependent methods that infer the confidence of early stopping from training data may be more suitable, with some likely trade-off in efficiency, especially in safety-critical domains.  
    \item \textbf{Dependency on initial PoT capability in bootstrapping:} Our case study on cross-modal bootstrapping assumes that the base model has at least some minimal capability in generating PoT solutions, even if weak.  While this is generally true for modern LLMs (all general-purpose models we are aware of exhibit at least minimal Python-like code generation ability due to code-rich pretraining data~\citep{jain2024livecodebench}), models with very limited or no prior exposure to programmatic reasoning may struggle to benefit from this bootstrapping process.
\end{itemize}

%% file: appendix.tex
\section{Appendix}
\label{sec:appendix}

\newcommand{\promptcaption}[2]{%
  \refstepcounter{figure}
  \label{#1}
  \textbf{Figure~\thefigure}:~#2\par\vspace{0.5em}%
}
\FloatBarrier 
\subsection{Prompts}
\label{sec:prompts}

This section contains the full prompts we used for chain-of-thought and program-of-thought inference. 


\begin{figure*}[t]
\centering
\begin{tcolorbox}[colback=white, colframe=black, title=CoT Prompt - GSM, left=1mm, fontupper=\footnotesize\ttfamily, breakable]
\begin{verbatim}
Please solve the given mathematical problem, doing step by step reasoning to arrive 
at the final answer. Please mark the final answer in a "\\boxed{}" annotation as 
shown in the example below.
------
Problem:
Let \\[f(x) = \\left\\{\n\\begin{array}{cl} ax+3, &\\text{ if }x>2, \\\\\nx-5
&\\text{ if } -2 \\le x \\le 2, \\\\\n2x-b &\\text{ if } x <-2.\n\\end{array}
\n\\right.\\]
Find $a+b$ if the piecewise function is continuous (which means
that its graph can be drawn without lifting your pencil from the paper).
Solution:
For the piecewise function to be continuous, the cases must "meet" at $2$ and $-2$. 
For example, $ax+3$ and $x-5$ must be equal when $x=2$. This implies $a(2)+3=2-5$, 
which we solve to get $2a=-6 \\Rightarrow a=-3$. Similarly, $x-5$ and $2x-b$ must 
be equal when $x=-2$. Substituting, we get $-2-5=2(-2)-b$, which implies $b=3$. 
So $a+b=-3+3=\\boxed{0}$.
------
Problem:
Jame's buys 100 head of cattle for $40,000.  It cost 20% more than that to feed them.  
They each weigh 1000 pounds and sell for $2 per pound.  
How much profit did he make?
Solution:
It cost .2*40000=$<<.2*40000=8000>>8000 more to feed than buy them
\nSo the total cost to feed them was 40000+8000=$<<40000+8000=48000>>48,000
\nSo in total it cost 48000+40000=$<<48000+40000=88000>>88,000
\nEach cow sells for 2*1000=$<<2*1000=2000>>2000\nSo he sells them for
2000*100=$<<2000*100=200000>>200,000
\nSo he makes a profit of 200,000-88,000=$<<200000-88000=112000>>112,000
\\\\boxed{112000}
------
Problem:
[[PROBLEM]]
\end{verbatim}
\end{tcolorbox}
\promptcaption{fig:cot-gms}{CoT prompt used for GSM8K, MATH, and SVAMP
(two in-domain demonstrations).}
\end{figure*}

\begin{figure*}[t]
\centering
\begin{tcolorbox}[colback=white, colframe=black, title=PoT Prompt - GSM, left=1mm, fontupper=\footnotesize\ttfamily]
\begin{verbatim}
Please generate Python code to solve the given mathematical problem. The code should 
store the final answer in a variable named 'ans' as shown in the example below.
------
Question:
Let \\[f(x) = \\left\\{\n\\begin{array}{cl} ax+3, &\\text{ if }x>2, 
\\\\\nx-5 &\\text{ if } -2 \\le x \\le 2, \\\\\n2x-b &\\text{ if } x <-2.
\n\\end{array}\n\\right.\\]
Find $a+b$ if the piecewise function is continuous (which means that its graph can 
be drawn without lifting your pencil from the paper).
PythonCode:
from sympy import symbols, Eq, solve

a, b = symbols('a b')
eqs = [
    Eq(2 * a + 3, -3),   # Match limits at x = 2
    Eq(-4 - b, -7)       # Match limits at x = -2
]
sol = solve(eqs, (a, b))
a_val, b_val = sol[a], sol[b]
ans = a_val + b_val
------
Question:
Jame's buys 100 head of cattle for $40,000.  It cost 20% more than that to feed them.  
They each weigh 1000 pounds and sell for $2 per pound.  How much profit did he make?
PythonCode:
num_cattle = 100
purchase_price = 40_000
feed_cost = purchase_price * 1.20
weight_per_cow = 1_000
price_per_pound = 2
revenue = num_cattle * weight_per_cow * price_per_pound
total_cost = purchase_price + feed_cost
ans = revenue - total_cost
------
Question:
[[QUESTION]]
PythonCode:
\end{verbatim}
\end{tcolorbox}
\promptcaption{fig:pot-gms}{PoT prompt used for GSM8K, MATH, and SVAMP
(two in-domain demonstrations).}
\end{figure*}


\begin{figure*}[t]
\centering
\begin{tcolorbox}[colback=white, colframe=black, title=CoT Prompt - FinQA, left=1mm, fontupper=\footnotesize\ttfamily]
\scriptsize
\begin{verbatim}

Please solve the given mathematical problem, doing step by step reasoning to arrive at
the final answer. Please mark the final answer in a "\\boxed{}" annotation. Be mindful of
units handling in your solution.  
------ 
Problem: 
for uncoated freesheet paper and market pulp announced at the end of 2009 become
effective . input costs are expected to be higher due to wood supply constraints at the kwidzyn
mill and annual tariff increases on energy in russia . planned main- tenance outage costs are
expected to be about flat , while operating costs should be favorable . asian printing papers net
sales were approx- imately $ 50 million in 2009 compared with approx- imately $ 20 million in
both 2008 and 2007 . operating earnings increased slightly in 2009 compared with 2008 , but
were less than $ 1 million in all periods . u.s . market pulp net sales in 2009 totaled $ 575
million compared with $ 750 million in 2008 and $ 655 million in 2007 . operating earnings in
2009 were $ 140 million ( a loss of $ 71 million excluding alter- native fuel mixture credits
and plant closure costs ) compared with a loss of $ 156 million ( a loss of $ 33 million
excluding costs associated with the perma- nent shutdown of the bastrop mill ) in 2008 and earn-
ings of $ 78 million in 2007 . sales volumes in 2009 decreased from 2008 levels due to weaker
global demand . average sales price realizations were significantly lower as the decline in
demand resulted in significant price declines for market pulp and smaller declines in fluff
pulp . input costs for wood , energy and chemicals decreased , and freight costs were
significantly lower . mill operating costs were favorable across all mills , and planned maintenance
downtime costs were lower . lack-of-order downtime in 2009 increased to approx- imately 540000
tons , including 480000 tons related to the permanent shutdown of our bastrop mill in the
fourth quarter of 2008 , compared with 135000 tons in 2008 . in the first quarter of 2010 ,
sales volumes are expected to increase slightly , reflecting improving customer demand for
fluff pulp , offset by slightly seasonally weaker demand for softwood and hard- wood pulp in
china . average sales price realizations are expected to improve , reflecting the realization
of previously announced sales price increases for fluff pulp , hardwood pulp and softwood
pulp . input costs are expected to increase for wood , energy and chemicals , and freight
costs may also increase . planned maintenance downtime costs will be higher , but operating
costs should be about flat . consumer packaging demand and pricing for consumer packaging
prod- ucts correlate closely with consumer spending and general economic activity . in
addition to prices and volumes , major factors affecting the profitability of consumer packaging
are raw material and energy costs , freight costs , manufacturing efficiency and product mix
. consumer packaging net sales in 2009 decreased 4% ( 4 % ) compared with 2008 and
increased 1% ( 1 % ) compared with 2007 . operating profits increased significantly compared with
both 2008 and 2007 . excluding alternative fuel mixture credits and facility closure costs ,
2009 operating profits were sig- nificantly higher than 2008 and 57% ( 57 % ) higher than
2007 . benefits from higher average sales price realizations ( $ 114 million ) , lower raw
material and energy costs ( $ 114 million ) , lower freight costs ( $ 21 million ) , lower costs
associated with the reorganiza- tion of the shorewood business ( $ 23 million ) , favor- able
foreign exchange effects ( $ 14 million ) and other items ( $ 12 million ) were partially offset
by lower sales volumes and increased lack-of-order downtime ( $ 145 million ) and costs
associated with the perma- nent shutdown of the franklin mill ( $ 67 million ) . additionally ,
operating profits in 2009 included $ 330 million of alternative fuel mixture credits . consumer
packaging in millions 2009 2008 2007 .
----------------------------------------------------------------
\n| in millions      | 2009   | 2008   | 2007  |
\n|------------------|--------|--------|--------|
\n| sales            | $ 3060 | $ 3195 | $ 3015 |
\n| operating profit | 433    | 17     |112    |
\n---------------------------------------------------------------- 
sales: 2009: $3060, 2008: $ 3195, 2007: $ 3015. 
operating profit: 2009: 433, 2008: 17, 2007: 112. 
north american consumer packaging net sales were $ 2.2 billion compared with $ 2.5 billion in 2008 and
$ 2.4 billion in 2007 . operating earnings in 2009 were $ 343 million ( $ 87 million
excluding alter- native fuel mixture credits and facility closure costs ) compared with $ 8
million ( $ 38 million excluding facility closure costs ) in 2008 and $ 70 million in 2007 .
coated paperboard sales volumes were lower in 2009 compared with 2008 reflecting weaker market
conditions . average sales price realizations were significantly higher , reflecting the full-year
realization of price increases implemented in the second half of 2008 . raw material costs for wood
, energy and chemicals were significantly lower in 2009 , while freight costs were also
favorable . operating costs , however , were unfavorable and planned main- tenance downtime costs
were higher . lack-of-order downtime increased to 300000 tons in 2009 from 15000 tons in 2008
due to weak demand . operating results in 2009 include income of $ 330 million for
alternative fuel mixture credits and $ 67 million of expenses for shutdown costs for the franklin
mill . foodservice sales volumes were lower in 2009 than in 2008 due to generally weak
\end{verbatim}
\end{tcolorbox}
\end{figure*}

\begin{figure*}[t]
\centering
\begin{tcolorbox}[colback=white, colframe=black, left=1mm, fontupper=\footnotesize\ttfamily]
\scriptsize
\begin{verbatim}
world-wide economic conditions . average sales price realizations were . what is the average value
for sales?  

Solution: 
To determine the average value for sales across the three years
provided (2009: $3060M, 2008: $3195M, 2007: $3015M), sum them and divide by 3: (3060 + 3195 +
3015) / 3 = 9270 / 3 = 3090. The final answer is \(\boxed{3090}\).  
------ 
Problem: 
interest rate cash flow hedges 2013 we report changes in the fair value of cash flow hedges in
accumulated other comprehensive loss until the hedged item affects earnings . at both december 31 ,
2008 and 2007 , we had reductions of $ 4 million recorded as an accumulated other
comprehensive loss that is being amortized on a straight-line basis through september 30 , 2014 . as
of december 31 , 2008 and 2007 , we had no interest rate cash flow hedges outstanding .
earnings impact 2013 our use of derivative financial instruments had the following impact on
pre-tax income for the years ended december 31 : millions of dollars 2008 2007 2006 .
--------------------------------------------------------------------------------------------
\n| millions of dollars                                               |2008 | 2007  | 2006    |
\n|-------------------------------------------------------------------|----|--------|---------|
\n| ( increase ) /decrease in interest expense from interest rate hedging |$ 1  | $ -8 (8) | $ -8 ( 8 ) |
\n| ( increase ) /decrease in fuel expense from fuel derivatives      | 1  |-1 ( 1 ) | 3       |
\n| increase/ ( decrease ) in pre-tax income                              | $ 2| $ -9 (9) | $ -5 (5)|
\n------------------------------------------------------------------------------------------
( increase ) /decrease in interest expense from interest rate hedging: 2008: $ 1, 2007:
$ -8 ( 8 ), 2006: $ -8 ( 8 ). ( increase ) /decrease in fuel expense from fuel
derivatives: 2008: 1, 2007: -1 ( 1 ), 2006: 3. increase/ ( decrease ) in pre-tax income: 2008: $ 2,
2007: $ -9 ( 9 ), 2006: $ -5 ( 5 ). fair value of debt instruments 2013 the fair value of our
short- and long-term debt was estimated using quoted market prices , where available , or
current borrowing rates . at december 31 , 2008 , the fair value of total debt is approximately
$ 247 million less than the carrying value . at december 31 , 2007 , the fair value of
total debt exceeded the carrying value by approximately $ 96 million . at december 31 , 2008
and 2007 , approximately $ 320 million and $ 181 million , respectively , of fixed-rate debt
securities contained call provisions that allowed us to retire the debt instruments prior to final
maturity , with the payment of fixed call premiums , or in certain cases , at par . sale of
receivables 2013 the railroad transfers most of its accounts receivable to union pacific receivables
, inc . ( upri ) , a bankruptcy-remote subsidiary , as part of a sale of receivables
facility . upri sells , without recourse on a 364-day revolving basis , an undivided interest in
such accounts receivable to investors . the total capacity to sell undivided interests to
investors under the facility was $ 700 million and $ 600 million at december 31 , 2008 and 2007 ,
respectively . the value of the outstanding undivided interest held by investors under the facility
was $ 584 million and $ 600 million at december 31 , 2008 and 2007 , respectively . upri
reduced the outstanding undivided interest held by investors due to a decrease in available
receivables at december 31 , 2008 . the value of the outstanding undivided interest held by
investors is not included in our consolidated financial statements . the value of the undivided
interest held by investors was supported by $ 1015 million and $ 1071 million of accounts
receivable held by upri at december 31 , 2008 and 2007 , respectively . at december 31 , 2008 and
2007 , the value of the interest retained by upri was $ 431 million and $ 471 million ,
respectively . this retained interest is included in accounts receivable in our consolidated
financial statements . the interest sold to investors is sold at carrying value , which
approximates fair value , and there is no gain or loss recognized from the transaction . the value of
the outstanding undivided interest held by investors could fluctuate based upon the
availability of eligible receivables and is directly affected by changing business volumes and credit
risks , including default and dilution . if default or dilution percentages were to increase
one percentage point , the amount of eligible receivables would decrease by $ 6 million .
should our credit rating fall below investment grade , the value of the outstanding undivided
interest held by investors would be reduced , and , in certain cases , the investors would have
the right to discontinue the facility . the railroad services the sold receivables ; however
, the railroad does not recognize any servicing asset or liability as the servicing fees
adequately compensate us for these responsibilities . the railroad collected approximately $ 17.8
billion and $ 16.1 billion during the years ended december 31 , 2008 and 2007 , respectively .
upri used certain of these proceeds to purchase new receivables under the facility. . what
was the difference in billions of sold receivables from 2007 to 2008?  

Solution: 
From the text, 2007 receivables sold were $16.1 billion while 2008’s were $17.8 billion. 
The difference is: 17.8 - 16.1 = 1.7. The final answer is \(\boxed{1.7}\).  
------ 
Problem: 
baker hughes ,age company notes to consolidated and combined financial statements bhge 2017 form 10-k |
\end{verbatim}
\end{tcolorbox}
\end{figure*}

\begin{figure*}[t]
\centering
\begin{tcolorbox}[colback=white, colframe=black, left=1mm, fontupper=\footnotesize\ttfamily]
\scriptsize
\begin{verbatim}
83 issuance pursuant to awards granted under the lti plan over its term which expires on
the date of the annual meeting of the company in 2027 . a total of 53.7 million shares of
class a common stock are available for issuance as of december 31 , 2017 . as a result of the
acquisition of baker hughes , on july 3 , 2017 , each outstanding baker hughes stock option was
converted into an option to purchase a share of class a common stock in the company . consequently
, we issued 6.8 million stock options which are fully vested . each converted option is
subject to the same terms and conditions as applied to the original option , and the per share
exercise price of each converted option was reduced by $ 17.50 to reflect the per share amount of
the special dividend pursuant to the agreement associated with the transactions .
additionally , as a result of the acquisition of baker hughes , there were 1.7 million baker hughes
restricted stock units ( rsus ) that were converted to bhge rsus at a fair value of $ 40.18 .
stock-based compensation cost is measured at the date of grant based on the calculated fair value of
the award and is generally recognized on a straight-line basis over the vesting period of
the equity grant . the compensation cost is determined based on awards ultimately expected
to vest ; therefore , we have reduced the cost for estimated forfeitures based on
historical forfeiture rates . forfeitures are estimated at the time of grant and revised , if
necessary , in subsequent periods to reflect actual forfeitures . there were no stock-based
compensation costs capitalized as the amounts were not material . during the year ended december 31 ,
2017 , we issued 2.1 million rsus and 1.6 million stock options under the lti plan . these
rsus and stock options generally vest in equal amounts over a three-year vesting period
provided that the employee has remained continuously employed by the company through such vesting
date . stock based compensation expense was $ 37 million in 2017 . included in this amount is
$ 15 million of expense which relates to the acceleration of equity awards upon
termination of employment of baker hughes employees with change in control agreements , and are
included as part of \"merger and related costs\" in the consolidated and combined statements of
income ( loss ) . as bhge llc is a pass through entity , any tax benefit would be recognized by
its partners . due to its cumulative losses , bhge is unable to recognize a tax benefit on
its share of stock related expenses . stock options the fair value of each stock option
granted is estimated using the black-scholes option pricing model . the following table presents
the weighted average assumptions used in the option pricing model for options granted under
the lti plan . the expected life of the options represents the period of time the options
are expected to be outstanding . the expected life is based on a simple average of the
vesting term and original contractual term of the awards . the expected volatility is based on
the historical volatility of our five main competitors over a six year period . the
risk-free interest rate is based on the observed u.s . treasury yield curve in effect at the time
the options were granted . the dividend yield is based on a five year history of dividend
payouts in baker hughes. .
----------------------------------------------------------------------------
\n|                                                     | 2017             |
\n|-----------------------------------------------------|------------------|
\n| expected life ( years )                             | 6                |
\n| risk-free interest rate                             | 2.1% ( 2.1 % )   |
\n| volatility                                          | 36.4% ( 36.4 % ) |
\n| dividend yield                                      | 1.2% ( 1.2 % )   |
\n| weighted average fair value per share at grant date | $12.32           |
\n--------------------------------------------------------------------------
expected life ( years ): 2017: 6. risk-free interest rate: 2017: 2.1% ( 2.1 % ).
volatility: 2017: 36.4% ( 36.4 % ). dividend yield: 2017: 1.2% ( 1.2 % ). weighted average fair
value per share at grant date: 2017: $ 12.32. what is the total value of rsus converted to
bhge rsus , in millions?  

Solution: 
From the text, 1.7 million RSUs were converted at $40.18 each, 
so 1.7 × 40.18 = 68.306 → approximately 68.3. The final answer is \(\boxed{68.3}\).
------ 
Problem: 
for uncoated freesheet paper and market pulp announced at the end of 2009
become effective . input costs are expected to be higher due to wood supply constraints at the
kwidzyn mill and annual tariff increases on energy in russia . planned main- tenance outage
costs are expected to be about flat , while operating costs should be favorable . asian
printing papers net sales were approx- imately $ 50 million in 2009 compared with approx- imately
$ 20 million in both 2008 and 2007 . operating earnings increased slightly in 2009
\end{verbatim}
\end{tcolorbox}
\end{figure*}
\begin{figure*}[t]
\centering
\begin{tcolorbox}[colback=white, colframe=black, left=1mm, fontupper=\footnotesize\ttfamily]
\scriptsize
\begin{verbatim}
compared with 2008 , but were less than $ 1 million in all periods . u.s . market pulp net sales
in 2009 totaled $ 575 million compared with $ 750 million in 2008 and $ 655 million in 2007
. operating earnings in 2009 were $ 140 million ( a loss of $ 71 million excluding alter-
native fuel mixture credits and plant closure costs ) compared with a loss of $ 156 million ( a
loss of $ 33 million excluding costs associated with the perma- nent shutdown of the bastrop
mill ) in 2008 and earn- ings of $ 78 million in 2007 . sales volumes in 2009 decreased from
2008 levels due to weaker global demand . average sales price realizations were significantly
lower as the decline in demand resulted in significant price declines for market pulp and
smaller declines in fluff pulp . input costs for wood , energy and chemicals decreased , and
freight costs were significantly lower . mill operating costs were favorable across all mills ,
and planned maintenance downtime costs were lower . lack-of-order downtime in 2009 increased
to approx- imately 540000 tons , including 480000 tons related to the permanent shutdown of
our bastrop mill in the fourth quarter of 2008 , compared with 135000 tons in 2008 . in the
first quarter of 2010 , sales volumes are expected to increase slightly , reflecting improving
customer demand for fluff pulp , offset by slightly seasonally weaker demand for softwood and
hard- wood pulp in china . average sales price realizations are expected to improve ,
reflecting the realization of previously announced sales price increases for fluff pulp , hardwood
pulp and softwood pulp . input costs are expected to increase for wood , energy and chemicals
, and freight costs may also increase . planned maintenance downtime costs will be higher
, but operating costs should be about flat . consumer packaging demand and pricing for
consumer packaging prod- ucts correlate closely with consumer spending and general economic
activity . in addition to prices and volumes , major factors affecting the profitability of
consumer packaging are raw material and energy costs , freight costs , manufacturing efficiency
and product mix . consumer packaging net sales in 2009 decreased 4% ( 4 % ) compared with
2008 and increased 1% ( 1 % ) compared with 2007 . operating profits increased significantly
compared with both 2008 and 2007 . excluding alternative fuel mixture credits and facility
closure costs , 2009 operating profits were sig- nificantly higher than 2008 and 57% ( 57 % )
higher than 2007 . benefits from higher average sales price realizations ( $ 114 million ) ,
lower raw material and energy costs ( $ 114 million ) , lower freight costs ( $ 21 million ) ,
lower costs associated with the reorganiza- tion of the shorewood business ( $ 23 million ) ,
favor- able foreign exchange effects ( $ 14 million ) and other items ( $ 12 million ) were
partially offset by lower sales volumes and increased lack-of-order downtime ( $ 145 million ) and
costs associated with the perma- nent shutdown of the franklin mill ( $ 67 million ) .
additionally , operating profits in 2009 included $ 330 million of alternative fuel mixture credits .
consumer packaging in millions 2009 2008 2007 .
----------------------------------------------------------------
\n| in millions      | 2009   | 2008   | 2007  |
\n|------------------|--------|--------|--------|
\n| sales            | $ 3060 | $ 3195 | $ 3015 |
\n| operating profit | 433    | 17     | 112    |
\n---------------------------------------------------------------- 
sales: 2009: $3060, 2008: $ 3195, 2007: $ 3015. 
operating profit: 2009: 433, 2008: 17, 2007: 112. 
north american consumer packaging net sales were $ 2.2 billion compared with $ 2.5 billion in 2008 and
$ 2.4 billion in 2007 . operating earnings in 2009 were $ 343 million ( $ 87 million
excluding alter- native fuel mixture credits and facility closure costs ) compared with $ 8
million ( $ 38 million excluding facility closure costs ) in 2008 and $ 70 million in 2007 .
coated paperboard sales volumes were lower in 2009 compared with 2008 reflecting weaker market
conditions . average sales price realizations were significantly higher , reflecting the full-year
realization of price increases implemented in the second half of 2008 . raw material costs for wood
, energy and chemicals were significantly lower in 2009 , while freight costs were also
favorable . operating costs , however , were unfavorable and planned main- tenance downtime costs
were higher . lack-of-order downtime increased to 300000 tons in 2009 from 15000 tons in 2008
due to weak demand . operating results in 2009 include income of $ 330 million for
alternative fuel mixture credits and $ 67 million of expenses for shutdown costs for the franklin
mill . foodservice sales volumes were lower in 2009 than in 2008 due to generally weak
world-wide economic conditions . average sales price realizations were . considering the years 2008
and 2009 , what is the variation observed in the operating profit , in millions?  

Solution:
The text provides operating profits for 2008 (17 million) and 2009 (433 million).
Subtracting the 2008 amount from the 2009 amount: 433 - 17 = 416. The final answer is
\(\boxed{416}\). 
------ 
Problem: 
[[PROBLEM]]       
\end{verbatim}
\end{tcolorbox}
\promptcaption{fig:cot-finqa}{CoT prompt used for FinQA
(four in-domain demonstrations).}
\end{figure*}

\begin{figure*}[t]
\centering
\begin{tcolorbox}[colback=white, colframe=black, title=PoT Prompt - FinQA, left=1mm, fontupper=\footnotesize\ttfamily]
\scriptsize
\begin{verbatim}
Please generate Python code to solve the given mathematical problem. The code should
first  define the input parameter values as stated in the problem, then provide the solution
code, and then store the final answer in a variable named 'ans'. Be mindful of units handling
in your solution. 
------ 
Question: for uncoated freesheet paper and market pulp announced
at the end of 2009 become effective . input costs are expected to be higher due to wood
supply constraints at the kwidzyn mill and annual tariff increases on energy in russia .
planned main- tenance outage costs are expected to be about flat , while operating costs should
be favorable . asian printing papers net sales were approx- imately $ 50 million in 2009
compared with approx- imately $ 20 million in both 2008 and 2007 . operating earnings increased
slightly in 2009 compared with 2008 , but were less than $ 1 million in all periods . u.s .
market pulp net sales in 2009 totaled $ 575 million compared with $ 750 million in 2008 and $
655 million in 2007 . operating earnings in 2009 were $ 140 million ( a loss of $ 71 million
excluding alter- native fuel mixture credits and plant closure costs ) compared with a loss of $
156 million ( a loss of $ 33 million excluding costs associated with the perma- nent
shutdown of the bastrop mill ) in 2008 and earn- ings of $ 78 million in 2007 . sales volumes in
2009 decreased from 2008 levels due to weaker global demand . average sales price
realizations were significantly lower as the decline in demand resulted in significant price declines
for market pulp and smaller declines in fluff pulp . input costs for wood , energy and
chemicals decreased , and freight costs were significantly lower . mill operating costs were
favorable across all mills , and planned maintenance downtime costs were lower . lack-of-order
downtime in 2009 increased to approx- imately 540000 tons , including 480000 tons related to the
permanent shutdown of our bastrop mill in the fourth quarter of 2008 , compared with 135000 tons
in 2008 . in the first quarter of 2010 , sales volumes are expected to increase slightly ,
reflecting improving customer demand for fluff pulp , offset by slightly seasonally weaker demand
for softwood and hard- wood pulp in china . average sales price realizations are expected to
improve , reflecting the realization of previously announced sales price increases for fluff
pulp , hardwood pulp and softwood pulp . input costs are expected to increase for wood ,
energy and chemicals , and freight costs may also increase . planned maintenance downtime costs
will be higher , but operating costs should be about flat . consumer packaging demand and
pricing for consumer packaging prod- ucts correlate closely with consumer spending and general
economic activity . in addition to prices and volumes , major factors affecting the profitability
of consumer packaging are raw material and energy costs , freight costs , manufacturing
efficiency and product mix . consumer packaging net sales in 2009 decreased 4% ( 4 % ) compared
with 2008 and increased 1% ( 1 % ) compared with 2007 . operating profits increased
significantly compared with both 2008 and 2007 . excluding alternative fuel mixture credits and
facility closure costs , 2009 operating profits were sig- nificantly higher than 2008 and 57% (
57 % ) higher than 2007 . benefits from higher average sales price realizations ( $ 114
million ) , lower raw material and energy costs ( $ 114 million ) , lower freight costs ( $ 21
million ) , lower costs associated with the reorganiza- tion of the shorewood business ( $ 23
million ) , favor- able foreign exchange effects ( $ 14 million ) and other items ( $ 12 million
) were partially offset by lower sales volumes and increased lack-of-order downtime ( $
145 million ) and costs associated with the perma- nent shutdown of the franklin mill ( $ 67
million ) . additionally , operating profits in 2009 included $ 330 million of alternative fuel
mixture credits . consumer packaging in millions 2009 2008 2007 .
----------------------------------------------------------------
\n| in millions      | 2009   | 2008   | 2007  |
\n|------------------|--------|--------|--------|
\n| sales            | $ 3060 | $ 3195 | $ 3015 |
\n| operating profit | 433    | 17     | 112    |
\n---------------------------------------------------------------- 
sales: 2009: $3060, 2008: $ 3195, 2007: $ 3015. operating profit: 2009: 433, 2008: 17, 2007: 
112. north american consumer packaging net sales were $ 2.2 billion compared with $ 2.5 billion in 2008 
and $ 2.4 billion in 2007 . operating earnings in 2009 were $ 343 million ( $ 87 million
excluding alter- native fuel mixture credits and facility closure costs ) compared with $ 8
million ( $ 38 million excluding facility closure costs ) in 2008 and $ 70 million in 2007 .
coated paperboard sales volumes were lower in 2009 compared with 2008 reflecting weaker market
conditions . average sales price realizations were significantly higher , reflecting the full-year
realization of price increases implemented in the second half of 2008 . raw material costs for wood
, energy and chemicals were significantly lower in 2009 , while freight costs were also
favorable . operating costs , however , were unfavorable and planned main- tenance downtime costs
were higher . lack-of-order downtime increased to 300000 tons in 2009 from 15000 tons in 2008
due to weak demand . operating results in 2009 include income of $ 330 million for alternative 
fuel mixture credits and $ 67 million of expenses for shutdown costs for the  franklin mill . 
foodservice sales volumes were lower in 2009 than in 2008 due to generally weak world-wide 
economic conditions . average sales price realizations were . what is the average value for sales? 
\end{verbatim}
\end{tcolorbox}
\end{figure*}
\begin{figure*}[t]
\centering
\begin{tcolorbox}[colback=white, colframe=black, left=1mm, fontupper=\footnotesize\ttfamily]
\scriptsize
\begin{verbatim}
PythonCode: 
# input parameters 
sales_2009 = 3060 
sales_2008 = 3195 
sales_2007 = 3015  

# solution code 
ans = (sales_2009 + sales_2008 + sales_2007) / 3  
------
Question: 
interest rate cash flow hedges 2013 we report changes in the fair value of cash flow hedges in
accumulated other comprehensive loss until the hedged item affects earnings . at both december 31 ,
2008 and 2007 , we had reductions of $ 4 million recorded as an accumulated other
comprehensive loss that is being amortized on a straight-line basis through september 30 , 2014 . as
of december 31 , 2008 and 2007 , we had no interest rate cash flow hedges outstanding .
earnings impact 2013 our use of derivative financial instruments had the following impact on
pre-tax income for the years ended december 31 : millions of dollars 2008 2007 2006 .
--------------------------------------------------------------------------------------------
\n| millions of dollars                                               |2008 | 2007  | 2006    |
\n|-------------------------------------------------------------------|----|--------|---------|
\n| ( increase ) /decrease in interest expense from interest rate hedging |$ 1  | $ -8 (8) | $ -8 ( 8 ) |
\n| ( increase ) /decrease in fuel expense from fuel derivatives      | 1  |-1 ( 1 ) | 3       |
\n| increase/ ( decrease ) in pre-tax income                              | $ 2| $ -9 (9) | $ -5 (5)|
\n------------------------------------------------------------------------------------------
( increase ) /decrease in interest expense from interest rate hedging: 2008: $ 1, 2007:
$ -8 ( 8 ), 2006: $ -8 ( 8 ). ( increase ) /decrease in fuel expense from fuel
derivatives: 2008: 1, 2007: -1 ( 1 ), 2006: 3. increase/ ( decrease ) in pre-tax income: 2008: $ 2,
2007: $ -9 ( 9 ), 2006: $ -5 ( 5 ). fair value of debt instruments 2013 the fair value of our
short- and long-term debt was estimated using quoted market prices , where available , or
current borrowing rates . at december 31 , 2008 , the fair value of total debt is approximately
$ 247 million less than the carrying value . at december 31 , 2007 , the fair value of
total debt exceeded the carrying value by approximately $ 96 million . at december 31 , 2008
and 2007 , approximately $ 320 million and $ 181 million , respectively , of fixed-rate debt
securities contained call provisions that allowed us to retire the debt instruments prior to final
maturity , with the payment of fixed call premiums , or in certain cases , at par . sale of
receivables 2013 the railroad transfers most of its accounts receivable to union pacific receivables
, inc . ( upri ) , a bankruptcy-remote subsidiary , as part of a sale of receivables
facility . upri sells , without recourse on a 364-day revolving basis , an undivided interest in
such accounts receivable to investors . the total capacity to sell undivided interests to
investors under the facility was $ 700 million and $ 600 million at december 31 , 2008 and 2007 ,
respectively . the value of the outstanding undivided interest held by investors under the facility
was $ 584 million and $ 600 million at december 31 , 2008 and 2007 , respectively . upri
reduced the outstanding undivided interest held by investors due to a decrease in available
receivables at december 31 , 2008 . the value of the outstanding undivided interest held by
investors is not included in our consolidated financial statements . the value of the undivided
interest held by investors was supported by $ 1015 million and $ 1071 million of accounts
receivable held by upri at december 31 , 2008 and 2007 , respectively . at december 31 , 2008 and
2007 , the value of the interest retained by upri was $ 431 million and $ 471 million ,
respectively . this retained interest is included in accounts receivable in our consolidated
financial statements . the interest sold to investors is sold at carrying value , which
approximates fair value , and there is no gain or loss recognized from the transaction . the value of
the outstanding undivided interest held by investors could fluctuate based upon the
availability of eligible receivables and is directly affected by changing business volumes and credit
risks , including default and dilution . if default or dilution percentages were to increase
one percentage point , the amount of eligible receivables would decrease by $ 6 million .
should our credit rating fall below investment grade , the value of the outstanding undivided
interest held by investors would be reduced , and , in certain cases , the investors would have
the right to discontinue the facility . the railroad services the sold receivables ; however
, the railroad does not recognize any servicing asset or liability as the servicing fees
adequately compensate us for these responsibilities . the railroad collected approximately $ 17.8
billion and $ 16.1 billion during the years ended december 31 , 2008 and 2007 , respectively .
upri used certain of these proceeds to purchase new receivables under the facility. . what
was the difference in billions of sold receivables from 2007 to 2008? 

PythonCode: 
# input parameters 
receivables_2007 = 16.1  # billions of dollars 
receivables_2008 = 17.8  # billions of dollars  

# solution code 
ans = receivables_2008 - receivables_2007 
\end{verbatim}
\end{tcolorbox}
\end{figure*}

\begin{figure*}[t]
\centering
\begin{tcolorbox}[colback=white, colframe=black, left=1mm, fontupper=\footnotesize\ttfamily]
\scriptsize
\begin{verbatim} 
------
Question: 
baker hughes , age company notes to consolidated and combined financial statements bhge 2017 form 10-k 
| 83 issuance pursuant to awards granted under the lti plan over its term which expires on the date of
the annual meeting of the company in 2027 . a total of 53.7 million shares of class a common
stock are available for issuance as of december 31 , 2017 . as a result of the acquisition of
baker hughes , on july 3 , 2017 , each outstanding baker hughes stock option was converted
into an option to purchase a share of class a common stock in the company . consequently , we
issued 6.8 million stock options which are fully vested . each converted option is subject to
the same terms and conditions as applied to the original option , and the per share exercise
price of each converted option was reduced by $ 17.50 to reflect the per share amount of the
special dividend pursuant to the agreement associated with the transactions . additionally , as
a result of the acquisition of baker hughes , there were 1.7 million baker hughes
restricted stock units ( rsus ) that were converted to bhge rsus at a fair value of $ 40.18 .
stock-based compensation cost is measured at the date of grant based on the calculated fair value of
the award and is generally recognized on a straight-line basis over the vesting period of
the equity grant . the compensation cost is determined based on awards ultimately expected
to vest ; therefore , we have reduced the cost for estimated forfeitures based on
historical forfeiture rates . forfeitures are estimated at the time of grant and revised , if
necessary , in subsequent periods to reflect actual forfeitures . there were no stock-based
compensation costs capitalized as the amounts were not material . during the year ended december 31 ,
2017 , we issued 2.1 million rsus and 1.6 million stock options under the lti plan . these
rsus and stock options generally vest in equal amounts over a three-year vesting period
provided that the employee has remained continuously employed by the company through such vesting
date . stock based compensation expense was $ 37 million in 2017 . included in this amount is
$ 15 million of expense which relates to the acceleration of equity awards upon
termination of employment of baker hughes employees with change in control agreements , and are
included as part of \"merger and related costs\" in the consolidated and combined statements of
income ( loss ) . as bhge llc is a pass through entity , any tax benefit would be recognized by
its partners . due to its cumulative losses , bhge is unable to recognize a tax benefit on
its share of stock related expenses . stock options the fair value of each stock option
granted is estimated using the black-scholes option pricing model . the following table presents
the weighted average assumptions used in the option pricing model for options granted under
the lti plan . the expected life of the options represents the period of time the options
are expected to be outstanding . the expected life is based on a simple average of the
vesting term and original contractual term of the awards . the expected volatility is based on
the historical volatility of our five main competitors over a six year period . the
risk-free interest rate is based on the observed u.s . treasury yield curve in effect at the time
the options were granted . the dividend yield is based on a five year history of dividend
payouts in baker hughes. .
----------------------------------------------------------------------------
\n|                                                     | 2017             |
\n|-----------------------------------------------------|------------------|
\n| expected life ( years )                             | 6                |
\n| risk-free interest rate                             | 2.1% ( 2.1 % )   |
\n| volatility                                          | 36.4% ( 36.4 % ) |
\n| dividend yield                                      | 1.2% ( 1.2 % )   |
\n| weighted average fair value per share at grant date | $12.32           |
\n--------------------------------------------------------------------------
expected life ( years ): 2017: 6. risk-free interest rate: 2017: 2.1% ( 2.1 % ).
volatility: 2017: 36.4% ( 36.4 % ). dividend yield: 2017: 1.2% ( 1.2 % ). weighted average fair
value per share at grant date: 2017: $ 12.32. what is the total value of rsus converted to
bhge rsus , in millions? 

PythonCode: 
# input parameters 
num_rsus = 1.7      # in millions of units 
fair_value = 40.18  # in dollars per share  

# solution code
ans = num_rsus * fair_value  
------
Question: 
for uncoated freesheet paper and market pulp announced at the end of 2009 become
effective .  input costs are expected to be higher due to wood supply constraints at the kwidzyn
mill and annual tariff increases on energy in russia .planned main- tenance outage costs are
expected to be about flat , while operating costs should be favorable . asian printing papers net
sales were approx- imately $ 50 million in 2009 compared with approx- imately $ 20 million in
both 2008 and 2007 . operating earnings increased slightly in 2009 compared with 2008 , but
were less than $ 1 million in all periods . u.s . market pulp net sales in 2009 totaled $ 575
\end{verbatim}
\end{tcolorbox}
\end{figure*}

\begin{figure*}[t]
\centering
\begin{tcolorbox}[colback=white, colframe=black, left=1mm, fontupper=\footnotesize\ttfamily]
\scriptsize
\begin{verbatim}
million compared with $ 750 million in 2008 and $ 655 million in 2007 . operating earnings in
2009 were $ 140 million ( a loss of $ 71 million excluding alter- native fuel mixture credits
and plant closure costs ) compared with a loss of $ 156 million ( a loss of $ 33 million
excluding costs associated with the perma- nent shutdown of the bastrop mill ) in 2008 and earn-
ings of $ 78 million in 2007 . sales volumes in 2009 decreased from 2008 levels due to weaker
global demand . average sales price realizations were significantly lower as the decline in
demand resulted in significant price declines for market pulp and smaller declines in fluff
pulp . input costs for wood , energy and chemicals decreased , and freight costs were
significantly lower . mill operating costs were favorable across all mills , and planned maintenance
downtime costs were lower . lack-of-order downtime in 2009 increased to approx- imately 540000
tons , including 480000 tons related to the permanent shutdown of our bastrop mill in the
fourth quarter of 2008 , compared with 135000 tons in 2008 . in the first quarter of 2010 ,
sales volumes are expected to increase slightly , reflecting improving customer demand for
fluff pulp , offset by slightly seasonally weaker demand for softwood and hard- wood pulp in
china . average sales price realizations are expected to improve , reflecting the realization
of previously announced sales price increases for fluff pulp , hardwood pulp and softwood
pulp . input costs are expected to increase for wood , energy and chemicals , and freight
costs may also increase . planned maintenance downtime costs will be higher , but operating
costs should be about flat . consumer packaging demand and pricing for consumer packaging
prod- ucts correlate closely with consumer spending and general economic activity . in
addition to prices and volumes , major factors affecting the profitability of consumer packaging
are raw material and energy costs , freight costs , manufacturing efficiency and product mix
. consumer packaging net sales in 2009 decreased 4% ( 4 % ) compared with 2008 and
increased 1% ( 1 % ) compared with 2007 . operating profits increased significantly compared with
both 2008 and 2007 . excluding alternative fuel mixture credits and facility closure costs ,
2009 operating profits were sig- nificantly higher than 2008 and 57% ( 57 % ) higher than
2007 . benefits from higher average sales price realizations ( $ 114 million ) , lower raw
material and energy costs ( $ 114 million ) , lower freight costs ( $ 21 million ) , lower costs
associated with the reorganiza- tion of the shorewood business ( $ 23 million ) , favor- able
foreign exchange effects ( $ 14 million ) and other items ( $ 12 million ) were partially offset
by lower sales volumes and increased lack-of-order downtime ( $ 145 million ) and costs
associated with the perma- nent shutdown of the franklin mill ( $ 67 million ) . additionally ,
operating profits in 2009 included $ 330 million of alternative fuel mixture credits . consumer
packaging in millions 2009 2008 2007 .
----------------------------------------------------------------
\n| in millions      | 2009   | 2008   | 2007  |
\n|------------------|--------|--------|--------|
\n| sales            | $ 3060 | $ 3195 | $ 3015 |
\n| operating profit | 433    | 17     | 112    |
\n---------------------------------------------------------------- 
sales: 2009: $3060, 2008: $ 3195, 2007: $ 3015. operating profit: 2009: 433, 2008: 17, 2007: 112. 
north american consumer packaging net sales were $ 2.2 billion compared with $ 2.5 billion in 2008 and
$ 2.4 billion in 2007 . operating earnings in 2009 were $ 343 million ( $ 87 million
excluding alter- native fuel mixture credits and facility closure costs ) compared with $ 8
million ( $ 38 million excluding facility closure costs ) in 2008 and $ 70 million in 2007 .
coated paperboard sales volumes were lower in 2009 compared with 2008 reflecting weaker market
conditions . average sales price realizations were significantly higher , reflecting the full-year
realization of price increases implemented in the second half of 2008 . raw material costs for wood
, energy and chemicals were significantly lower in 2009 , while freight costs were also
favorable . operating costs , however , were unfavorable and planned main- tenance downtime costs
were higher . lack-of-order downtime increased to 300000 tons in 2009 from 15000 tons in 2008
due to weak demand . operating results in 2009 include income of $ 330 million for
alternative fuel mixture credits and $ 67 million of expenses for shutdown costs for the franklin
mill . foodservice sales volumes were lower in 2009 than in 2008 due to generally weak
world-wide economic conditions . average sales price realizations were . considering the years 2008
and 2009 , what is the variation observed in the operating profit , in millions? 

PythonCode:
# input parameters 
operating_profit_2009 = 433  # in millions of dollars
operating_profit_2008 = 17   # in millions of dollars  

# solution code
ans = operating_profit_2009 - operating_profit_2008  
------
Question: 
[[QUESTION]]
PythonCode:
\end{verbatim}
\end{tcolorbox}
\promptcaption{fig:pot-finqa}{PoT prompt used for FinQA
(four in-domain demonstrations).}
\end{figure*}


\begin{figure*}[t]
\centering
\begin{tcolorbox}[colback=white, colframe=black,
                  title=CoT Prompt - TabMWP, left=1mm, fontupper=\footnotesize\ttfamily]
\begin{verbatim}
Please solve the given mathematical problem, doing step by step reasoning 
to arrive at the final answer. Please mark the final answer in a "\\boxed{}" 
annotation as shown in the example below.
------
Problem:
A bus driver paid attention to how many passengers her bus had each day. 
On which day did the bus have the fewest passengers?

People on the bus:
Day | Number of people
Monday | 39
Tuesday | 38
Wednesday | 32
Thursday | 36

Solution:
Find the least number in the table. Remember to compare the numbers 
tarting with the highest place value. The least number is 32.

Now find the corresponding day. Wednesday corresponds to 32. 
The final answer is \boxed{Wednesday}.

------

Problem:
Jayla has $95.35. How much money will Jayla have left if she buys a CD player 
and a DVD?

None:
CD | $13.37
CD player | $16.61
DVD | $13.28
alarm clock | $13.72
microwave | $53.38

Solution:
Find the total cost of a CD player and a DVD.

$16.61 + $13.28 = $29.89

Now subtract the total cost from the tarting amount.

$95.35 - $29.89 = $65.46

Jayla will have $65.46 left. The final answer is \boxed{65.46 $}.
------
Problem:
[[PROBLEM]]
\end{verbatim}
\end{tcolorbox}
\promptcaption{fig:cot-tabmwp}{CoT prompt used for TabMWP
(two in-domain demonstrations).}
\end{figure*}

\begin{figure*}[t]
\centering
\begin{tcolorbox}[colback=white, colframe=black,
                  title=PoT Prompt - TabMWP, left=1mm, fontupper=\footnotesize\ttfamily]
\begin{verbatim}
Please generate Python code to solve the given mathematical problem. The 
code should store the final answer in a variable named 'ans' as shown in the 
example below.
------
Question:
A bus driver paid attention to how many passengers her bus had each day. On 
which  day did the bus have the fewest passengers?

People on the bus:
Day | Number of people
Monday | 39
Tuesday | 38
Wednesday | 32
Thursday | 36
PythonCode:
bus_data = {
    "Monday": 39,
    "Tuesday": 38,
    "Wednesday": 32,
    "Thursday": 36
}

min_passengers = float('inf')
ans = None

for day, passengers in bus_data.items():
    if passengers < min_passengers:
        min_passengers = passengers
        ans = day
------
Question:
Jayla has $95.35. How much money will Jayla have left if she buys a CD player 
and a DVD?

None:
CD | $13.37
CD player | $16.61
DVD | $13.28
alarm clock | $13.72
microwave | $53.38
PythonCode:
# Input parameters
cd_player_cost = 16.61
dvd_cost      = 13.28
initial_amount = 95.35

# Solution
ans = initial_amount - (cd_player_cost + dvd_cost)
------
Question:
[[QUESTION]]
PythonCode:
\end{verbatim}
\end{tcolorbox}
\promptcaption{fig:pot-tabmwp}{PoT prompt used for TabMWP
(two in-domain demonstrations).}
\end{figure*}

\clearpage

\begin{figure*}[t]
\centering
\begin{tcolorbox}[colback=white, colframe=black, title=CoT-to-PoT Generation Prompt for Boostrapping PoT Rationales, left=1mm, fontupper=\footnotesize\ttfamily]
\scriptsize
\begin{verbatim}
Please generate Python code to solve the given mathematical problem. The code should first define the 
input parameter values as stated in the problem, then provide the solution code, and then store the 
final answer in a variable named 'ans'. The first 8 question and python code pairs are given as an 
example, solve the last question.
------
Question:
There are 15 trees in the grove. Grove workers will plant trees in the grove today. After they are 
done, there will be 21 trees. How many trees did the grove workers plant today?
Solution:
There are 15 trees originally. Then there were 21 trees after some more were planted. So there must 
have been 21 - 15 = 6. The final answer is \\boxed{6}
PythonCode:
# input parameters
initial_trees = 15
final_trees = 21

# solution code
ans = final_trees - initial_trees
------
Question:
If there are 3 cars in the parking lot and 2 more cars arrive, how many cars are in the parking lot?
Solution:
There are originally 3 cars. 2 more cars arrive. 3 + 2 = 5. The final answer is \\boxed{5}
PythonCode:
# input parameters
initial_cars = 3
arrived_cars = 2

# solution code
ans = initial_cars + arrived_cars
------
Question:
Leah had 32 chocolates and her sister had 42. If they ate 35, how many pieces do they have left in 
total?
Solution:
Originally, Leah had 32 chocolates. Her sister had 42. So in total they had 32 + 42 = 74. After eating 
35, they had 74 - 35 = 39. The final answer is \\boxed{39}
PythonCode:
# input parameters
leah_chocolates = 32
sister_chocolates = 42
eaten_chocolates = 35

# solution code
total_chocolates = leah_chocolates + sister_chocolates
ans = total_chocolates - eaten_chocolates
------
Question:
Jason had 20 lollipops. He gave Denny some lollipops. Now Jason has 12 lollipops. How many lollipops 
did Jason give to Denny?
Solution:
Jason started with 20 lollipops. Then he had 12 after giving some to Denny. So he gave Denny 20 - 12 = 
8. The final answer is \\boxed{8}
PythonCode:
# input parameters
initial_lollipops = 20
remaining_lollipops = 12

# solution code
ans = initial_lollipops - remaining_lollipops
------
Question:
Shawn has five toys. For Christmas, he got two toys each from his mom and dad. How many toys does he 
have now?
Solution:
Shawn started with 5 toys. If he got 2 toys each from his mom and dad, then that is 4 more toys. 
5 + 4 = 9. The final answer is \\boxed{9}
\end{verbatim}
\end{tcolorbox}
\end{figure*}

\begin{figure*}[t]
\centering
\begin{tcolorbox}[colback=white, colframe=black, left=1mm, fontupper=\footnotesize\ttfamily]
\scriptsize
\begin{verbatim}
PythonCode:
# input parameters
initial_toys = 5
toys_from_mom = 2
toys_from_dad = 2

# solution code
ans = initial_toys + toys_from_mom + toys_from_dad
------
Question:
There were nine computers in the server room. Five more computers were installed each day, from Monday 
to thursday. How many computers are now in the server room?
Solution:
There were originally 9 computers. For each of 4 days, 5 more computers were added. So 5 * 4 = 20 
computers were added. 9 + 20 is 29. The final answer is \\boxed{29}
PythonCode:
# input parameters
initial_computers = 9
additional_computers_per_day = 5
days = 4

# solution code
total_additional_computers = additional_computers_per_day * days
ans = initial_computers + total_additional_computers
------
Question:
Michael had 58 golf balls. On tuesday, he lost 23 golf balls. On wednesday, he lost 2 more. How many 
golf balls did he have at the end of wednesday?
Solution:
Michael started with 58 golf balls. After losing 23 on tuesday, he had 58 - 23 = 35. After losing 2 
more, he had 35 - 2 = 33 golf balls. The final answer is \\boxed{33}
PythonCode:
# input parameters
initial_golf_balls = 58
lost_golf_balls_tuesday = 23
lost_golf_balls_wednesday = 2

# solution code
remaining_golf_balls = initial_golf_balls - lost_golf_balls_tuesday
ans = remaining_golf_balls - lost_golf_balls_wednesday
------
Question:
Olivia has $23. She bought five bagels for $3 each. How much money does she have left?
Solution:
Olivia had 23 dollars. 5 bagels for 3 dollars each will be 5 x 3 = 15 dollars. So she has 23 - 15 
dollars left. 23 - 15 is 8. The final answer is \\boxed{8}
PythonCode:
# input parameters
initial_money = 23
bagel_cost = 3
num_bagels = 5

# solution code
total_cost = bagel_cost * num_bagels
ans = initial_money - total_cost
------
Question:
[[QUESTION]]
Solution:
[[SOLUTION]]
PythonCode:
\end{verbatim}
\end{tcolorbox}
\promptcaption{fig:cot-to-pot}{Prompt used for generation of PoT rationales from CoT in Case Study.}
\label{appfig:cottopot}
\end{figure*}

\subsection{Full Table Ensemble Results}
\newcommand{\cpfullpot}{\methodname{SC\textsubscript{PoT}}}   

\subsubsection{Full-sampling results}
This section reports the complete numbers for all full-sampling
variants for each dataset and model separately.  

Tables \ref{tab:gpt35-full-sampling} - \ref{tab:deepseek-full-sampling}
give per-dataset accuracies for GPT-3.5, GPT-4o,
\textsc{Mistral-large}, Qwen3-Coder and DeepSeek-R1.  The final row reports the 
average across the five benchmarks.\footnote{%
  GSM8K, MATH, SVAMP, FinQA, and TabMWP.
}

\begin{table}[!t]
\centering
\small
\begin{tabular}{lccccc}
\toprule
\textbf{Dataset} &
\selfconsistency & \cpfullmaj & \cpfullmax & \cpfullagr & \cpfullpot \\
\midrule
\gsm         & 89.4 & \textbf{91.4} & 90.6 & 90.8 & 82.0 \\
\mathdataset & 52.6 & \textbf{56.4} & 56.2 & 55.4 & 43.6 \\
\svamp       & 91.4 & \textbf{93.1} & 92.4 & \textbf{93.1} & 89.0 \\
\finqa       & 57.0 & 60.0 & \textbf{60.2} & 60.0 & 58.0 \\
\tabwmp      & \textbf{78.8} & 77.6 & 77.2 & 77.6 & 72.8 \\
\midrule
\textbf{Average} & 73.8 & \textbf{75.7} & 75.3 & 75.4 & 69.1 \\
\bottomrule
\end{tabular}
\caption{Full-sampling accuracy (\%) on \textbf{GPT-3.5}.}
\label{tab:gpt35-full-sampling}

\bigskip
\centering
\small
\begin{tabular}{lccccc}
\toprule
\textbf{Dataset} &
\selfconsistency & \cpfullmaj & \cpfullmax & \cpfullagr & \cpfullpot \\
\midrule
\gsm         & \textbf{98.0} & 97.8 & \textbf{98.0} & 97.8 & 97.2 \\
\mathdataset & 75.8 & \textbf{77.8} & 77.6 & 77.2 & 69.0 \\
\svamp       & 94.8 & 96.2 & 96.2 & 96.2 & \textbf{96.6} \\
\finqa       & 73.2 & 73.0 & 73.2 & 72.8 & 72.6 \\
\tabwmp      & 89.8 & 91.0 & 91.0 & 91.0 & 88.4 \\
\midrule
\textbf{Average} & 86.3 & \textbf{87.2} & \textbf{87.2} & 87.0 & 84.8 \\
\bottomrule
\end{tabular}
\caption{Full-sampling accuracy (\%) on \textbf{GPT-4o}.}
\label{tab:gpt4o-full-sampling}

\bigskip
\centering
\small
\begin{tabular}{lccccc}
\toprule
\textbf{Dataset} &
\selfconsistency & \cpfullmaj & \cpfullmax & \cpfullagr & \cpfullpot \\
\midrule
\gsm         & 97.4 & 97.4 & 97.4 & \textbf{97.6} & 96.8 \\
\mathdataset & 71.2 & 74.2 & \textbf{74.6} & 74.0 & 66.2 \\
\svamp       & 93.8 & \textbf{94.8} & \textbf{94.8} & 94.5 & \textbf{94.8} \\
\finqa       & 72.6 & 73.6 & \textbf{74.0} & 73.6 & 73.0 \\
\tabwmp      & 89.8 & 90.6 & 90.6 & \textbf{90.8} & 89.2 \\
\midrule
\textbf{Average} & 85.0 & 86.1 & \textbf{86.3} & 86.1 & 84.0 \\
\bottomrule
\end{tabular}
\caption{Full-sampling accuracy (\%) on \textbf{Mistral-large}.}
\label{tab:mistral-full-sampling}

\bigskip
\centering
\small
\begin{tabular}{lccccc}
\toprule
\textbf{Dataset} &
\selfconsistency & \cpfullmaj & \cpfullmax & \cpfullagr & \cpfullpot \\
\midrule
\gsm         & \textbf{96.2} & \textbf{96.2} & \textbf{96.2} & \textbf{96.2} & \textbf{96.2} \\
\mathdataset & 85.2 & 84.8 & \textbf{85.4} & \textbf{85.4} & 78.8 \\
\svamp       & \textbf{96.6} & \textbf{96.6} & \textbf{96.6} & \textbf{96.6} & \textbf{96.6} \\
\finqa       & 69.2 & 72.2 & 72.2 & 72.2 & \textbf{72.8} \\
\tabwmp      & 91.4 & \textbf{91.8} & \textbf{91.8} & 91.6 & 90.8 \\
\midrule
\textbf{Average} & 87.7 & 88.3 & \textbf{88.4} & \textbf{88.4} & 87.0 \\
\bottomrule
\end{tabular}
\caption{Full-sampling accuracy (\%) on \textbf{Qwen3-Coder}.}
\label{tab:qwen-full-sampling}

\bigskip
\centering
\small
\begin{tabular}{lccccc}
\toprule
\textbf{Dataset} &
\selfconsistency & \cpfullmaj & \cpfullmax & \cpfullagr & \cpfullpot \\
\midrule
\gsm         & 97.2 & \textbf{97.6} & \textbf{97.6} & \textbf{97.6} & \textbf{97.6} \\
\mathdataset & \textbf{88.8} & \textbf{88.8} & \textbf{88.8} & \textbf{88.8} & 83.6 \\
\svamp       & 97.6 & \textbf{97.9} & \textbf{97.9} & \textbf{97.9} & 94.5 \\
\finqa       & 68.6 & 71.2 & 71.2 & 71.2 & \textbf{73.4} \\
\tabwmp      & \textbf{91.0} & \textbf{91.0} & \textbf{91.0} & 90.6 & 90.6 \\
\midrule
\textbf{Average} & 88.6 & \textbf{89.3} & \textbf{89.3} & 89.2 & 87.9 \\
\bottomrule
\end{tabular}
\caption{Full-sampling accuracy (\%) on \textbf{DeepSeek R1}.}
\label{tab:deepseek-full-sampling}
\end{table}

\subsubsection{Early-stopping results}
We provide full results  for early stopping for each dataset and model. Tables~\ref{tab:gpt35_early_stopping_results} - %
\ref{tab:deepseek_early_stopping_results} list accuracies and average
sample budgets for all models.%

\begin{table*}[t]
\centering
\small
\setlength{\tabcolsep}{5pt}
\begin{tabular}{l*{7}{c}|*{7}{c}}
\toprule
               & \multicolumn{7}{c}{\textbf{Accuracy (\%)}} &
                 \multicolumn{7}{c}{\textbf{Avg.\ \# Samples}} \\
\cmidrule(lr){2-8}\cmidrule(lr){9-15}
\textbf{Dataset} &
\adaptiveconsistency
& \adaptiveconsistencycotpot
& \cpearlyaa & \cpearlyfa & \cpearlyff &
  \cpearlybayesaa & \cpearlybayesfa \quad&\quad
\adaptiveconsistency
& \adaptiveconsistencycotpot
& \cpearlyaa & \cpearlyfa & \cpearlyff &
  \cpearlybayesaa & \cpearlybayesfa \\
\cmidrule(lr){1-1}\cmidrule(lr){2-8}\cmidrule(lr){9-15}
\gsm          & 89.6 & \textbf{91.6} & 89.8 & 90.6 & \textbf{91.6} & 90.8 & 90.6 \quad&\quad
               8.5 & 10.7 &  \textbf{4.2} &  5.3 &  8.4 &  4.9 &  6.9 \\
\mathdataset  & 50.0 & 54.2 & 55.4 & 53.4 & 54.4 & \textbf{55.8} & 53.8 \quad&\quad
              22.5 & 23.8 & \textbf{12.8} & 17.0 & 22.6 & 13.3 & 18.0 \\
\svamp        & 90.7 & \textbf{92.4} & 90.7 & 92.1 & \textbf{92.4} & 90.7 & 92.1 \quad&\quad
               7.9 &  7.7 &  \textbf{3.4} &  3.8 &  5.6 &  \textbf{3.4} &  4.6 \\
\finqa        & 68.4 & \textbf{70.6} & 69.2 & 69.4 & \textbf{70.6} & 69.2 & 70.0 \quad&\quad
               9.7 & 11.5 &  \textbf{5.4} &  6.1 &  9.8 &  \textbf{5.4} &  6.7 \\
\tabwmp       & \textbf{79.2} & 77.8 & 78.4 & 78.6 & 78.0 & 78.4 & 78.0 \quad&\quad
               9.8 & 10.1 &  \textbf{4.2} &  4.8 &  7.9 &  4.3 &  5.5 \\
\\[-0.55em]
\toprule
\textbf{Average} & 75.6 & 77.3 & 76.7 & 76.8 & \textbf{77.4} & 77.0 & 76.9 \quad&\quad
                 11.7 & 12.8 &  \textbf{6.0} &  7.4 & 10.9 &  6.2 &  8.4 \\
\bottomrule
\end{tabular}
\caption{Accuracy and average sample budget for \textbf{GPT-3.5}.}
\label{tab:gpt35_early_stopping_results}
\end{table*}

\begin{table*}[t]
\centering
\small
\setlength{\tabcolsep}{5pt}
\begin{tabular}{l*{7}{c}|*{7}{c}}
\toprule
               & \multicolumn{7}{c}{\textbf{Accuracy (\%)}} &
                 \multicolumn{7}{c}{\textbf{Avg.\ \# Samples}} \\
\cmidrule(lr){2-8}\cmidrule(lr){9-15}
\textbf{Dataset} &
\adaptiveconsistency
& \adaptiveconsistencycotpot
& \cpearlyaa & \cpearlyfa & \cpearlyff &
  \cpearlybayesaa & \cpearlybayesfa \quad&\quad
\adaptiveconsistency
& \adaptiveconsistencycotpot
& \cpearlyaa & \cpearlyfa & \cpearlyff &
  \cpearlybayesaa & \cpearlybayesfa \\
\cmidrule(lr){1-1}\cmidrule(lr){2-8}\cmidrule(lr){9-15}
\gsm          & \textbf{98.0} & \textbf{98.0} & 97.0 & 96.8 & 97.8 & 97.0 & 96.8 \quad&\quad
               4.6 &  4.8 &  \textbf{2.3} &  2.4 &  2.8 &  \textbf{2.3} &  2.4 \\
\mathdataset  & 75.6 & 75.8 & 76.8 & 76.6 & \textbf{78.0} & 76.8 & 77.0 \quad&\quad
              11.2 & 13.3 &  \textbf{7.2} &  8.7 & 11.5 &  \textbf{7.2} &  8.8 \\
\svamp        & 94.8 & 94.5 & \textbf{96.2} & \textbf{96.2} & \textbf{96.2} & \textbf{96.2} & \textbf{96.2} \quad&\quad
               4.8 &  4.9 &  \textbf{2.3} &  \textbf{2.3} &  2.7 &  \textbf{2.3} &  \textbf{2.3} \\
\finqa        & \textbf{73.2} & \textbf{73.2} & 72.8 & \textbf{73.2} & 73.0 & 72.8 & 73.0 \quad&\quad
               5.9 &  6.8 &  \textbf{3.2} &  3.3 &  4.9 &  \textbf{3.2} &  3.7 \\
\tabwmp       & 89.8 & 90.4 & \textbf{91.0} & \textbf{91.0} & \textbf{91.0} & \textbf{91.0} & \textbf{91.0} \quad&\quad
               5.2 &  6.9 &  \textbf{4.7} &  \textbf{4.7} &  5.1 &  \textbf{4.7} &  \textbf{4.7} \\
\\[-0.55em]
\toprule
\textbf{Average} & 86.3 & 86.4 & 86.8 & 86.8 & \textbf{87.2} & 86.8 & 86.8 \quad&\quad
                  6.3 &  7.3 &  \textbf{3.9} &  4.3 &  5.4 &  \textbf{3.9} &  4.4 \\
\bottomrule
\end{tabular}
\caption{Accuracy and average sample budget for \textbf{GPT-4o}.}
\label{tab:gpt4o_early_stopping_results}
\end{table*}

\newpage

\begin{table*}[t]
\centering
\small
\setlength{\tabcolsep}{5pt}
\begin{tabular}{l*{7}{c}|*{7}{c}}
\toprule
                 & \multicolumn{7}{c}{\textbf{Accuracy (\%)}} &
                   \multicolumn{7}{c}{\textbf{Avg.\ \# Samples}} \\
\cmidrule(lr){2-8}\cmidrule(lr){9-15}
\textbf{Dataset} &
\adaptiveconsistency & \adaptiveconsistencycotpot & \cpearlyaa & \cpearlyfa & \cpearlyff &
\cpearlybayesaa & \cpearlybayesfa \quad&\quad
\adaptiveconsistency & \adaptiveconsistencycotpot & \cpearlyaa & \cpearlyfa & \cpearlyff &
\cpearlybayesaa & \cpearlybayesfa \\
\cmidrule(lr){1-1}\cmidrule(lr){2-8}\cmidrule(lr){9-15}
\gsm          & 97.6 & \textbf{98.0} & \textbf{98.0} & \textbf{98.0} & 97.8 & \textbf{98.0} & \textbf{98.0} \quad&\quad
               4.5 &  4.7 &  \textbf{2.2} &  \textbf{2.2} &  2.4 &  \textbf{2.2} &  \textbf{2.2} \\
\mathdataset  & 71.2 & 70.8 & 73.4 & 73.6 & 74.0 & \textbf{74.4} & 74.0 \quad&\quad
              12.6 & 14.5 &  \textbf{7.4} &  9.2 & 12.7 &  7.8 & 10.7 \\
\svamp        & 93.8 & 93.8 & 93.8 & 93.8 & \textbf{94.8} & 93.8 & \textbf{94.8} \quad&\quad
               5.0 &  5.1 &  \textbf{2.2} &  2.5 &  3.0 &  \textbf{2.2} &  2.6 \\
\finqa        & 72.6 & 72.4 & \textbf{74.0} & 73.8 & 73.2 & 73.8 & 73.8 \quad&\quad
               6.6 &  7.6 &  \textbf{3.3} &  3.7 &  5.2 &  \textbf{3.3} &  4.3 \\
\tabwmp       & 90.0 & 90.0 & \textbf{90.2} & \textbf{90.2} & \textbf{90.2} & \textbf{90.2} & \textbf{90.2} \quad&\quad
               5.0 &  7.0 &  \textbf{4.3} &  4.4 &  5.0 &  \textbf{4.3} &  4.4 \\
\\[-0.55em]
\toprule
\textbf{Average} & 85.0 & 85.0 & 85.9 & 85.9 & 86.0 & 86.0 & \textbf{86.2} \quad&\quad
                  6.7 &  7.8 &  \textbf{3.9} &  4.4 &  5.7 &  4.0 &  4.8 \\
\bottomrule
\end{tabular}
\caption{Accuracy and average sample budget for \textbf{Mistral-large}.}
\label{tab:mistral_early_stopping_results}
\end{table*}

\begin{table*}[t]
\centering
\small
\setlength{\tabcolsep}{5pt}
\begin{tabular}{l*{7}{c}|*{7}{c}}
\toprule
               & \multicolumn{7}{c}{\textbf{Accuracy (\%)}} &
                 \multicolumn{7}{c}{\textbf{Avg.\ \# Samples}} \\
\cmidrule(lr){2-8}\cmidrule(lr){9-15}
\textbf{Dataset} &
\adaptiveconsistency & \adaptiveconsistencycotpot & \cpearlyaa & \cpearlyfa & \cpearlyff &
\cpearlybayesaa & \cpearlybayesfa \quad&\quad
\adaptiveconsistency & \adaptiveconsistencycotpot & \cpearlyaa & \cpearlyfa & \cpearlyff &
\cpearlybayesaa & \cpearlybayesfa \\
\cmidrule(lr){1-1}\cmidrule(lr){2-8}\cmidrule(lr){9-15}
\finqa        & 69.2 & 72.2 & 72.0 & 72.2 & 72.0 & 72.0 & 71.8 \quad&\quad
                6.2  &  8.3 &  4.2 &  4.3 &  6.2 &  4.3 &  5.9 \\
\tabwmp       & 91.4 & 91.8 & 91.4 & 91.6 & 91.8 & 91.4 & 91.6 \quad&\quad
                5.0  &  6.5 &  4.0 &  4.2 &  4.7 &  4.0 &  4.2 \\
\svamp        & 96.6 & 96.6 & 96.6 & 96.6 & 96.6 & 96.6 & 96.6 \quad&\quad
                4.3  &  4.2 &  2.2 &  2.2 &  2.2 &  2.2 &  2.2 \\
\gsm          & 96.0 & 96.2 & 96.6 & 96.6 & 96.4 & 96.6 & 96.4 \quad&\quad
                4.7  &  4.9 &  2.3 &  2.3 &  2.8 &  2.3 &  2.4 \\
\mathdataset  & 85.4 & 84.8 & 85.2 & 85.4 & 84.8 & 85.2 & 85.4 \quad&\quad
                9.1  & 10.4 &  6.4 &  7.0 &  8.7 &  6.4 &  7.0 \\
\\[-0.55em]
\toprule
\textbf{Average} & 87.7 & 88.3 & 88.4 & 88.5 & 88.3 & 88.4 & 88.4 \quad&\quad
                   5.9 &  6.9 &  3.8 &  4.0 &  4.9 &  3.8 &  4.3 \\
\bottomrule
\end{tabular}
\caption{Accuracy and average sample budget for \textbf{Qwen3-Coder}.}
\label{tab:qwen_early_stopping_results}
\end{table*}

\begin{table*}[t]
\centering
\small
\setlength{\tabcolsep}{5pt}
\begin{tabular}{l*{7}{c}|*{7}{c}}
\toprule
               & \multicolumn{7}{c}{\textbf{Accuracy (\%)}} &
                 \multicolumn{7}{c}{\textbf{Avg.\ \# Samples}} \\
\cmidrule(lr){2-8}\cmidrule(lr){9-15}
\textbf{Dataset} &
\adaptiveconsistency & \adaptiveconsistencycotpot & \cpearlyaa & \cpearlyfa & \cpearlyff &
\cpearlybayesaa & \cpearlybayesfa \quad&\quad
\adaptiveconsistency & \adaptiveconsistencycotpot & \cpearlyaa & \cpearlyfa & \cpearlyff &
\cpearlybayesaa & \cpearlybayesfa \\
\cmidrule(lr){1-1}\cmidrule(lr){2-8}\cmidrule(lr){9-15}
\finqa        & 68.4 & 71.2 & 71.6 & 71.4 & 71.4 & 71.6 & 71.8 \quad&\quad
                6.1  &  6.9 &  3.4 &  3.4 &  4.6 &  3.4 &  3.6 \\
\tabwmp       & 91.0 & 91.0 & 90.6 & 90.8 & 90.8 & 90.6 & 90.8 \quad&\quad
                4.1  &  6.7 &  4.5 &  4.5 &  4.8 &  4.5 &  4.5 \\
\svamp        & 97.6 & 97.9 & 97.9 & 97.9 & 97.9 & 97.9 & 97.9 \quad&\quad
                4.3  &  5.2 &  2.4 &  2.4 &  3.0 &  2.4 &  2.4 \\
\gsm          & 97.2 & 97.6 & 97.2 & 97.2 & 97.6 & 97.2 & 97.2 \quad&\quad
                4.1  &  4.4 &  2.1 &  2.1 &  2.4 &  2.1 &  2.1 \\
\mathdataset  & 88.8 & 88.8 & 88.4 & 89.0 & 88.8 & 88.4 & 88.6 \quad&\quad
                7.9  &  9.3 &  6.6 &  6.9 &  7.5 &  6.6 &  7.3 \\
\\[-0.55em]
\toprule
\textbf{Average} & 88.6 & 89.3 & 89.1 & 89.3 & 89.3 & 89.1 & 89.3 \quad&\quad
                   5.3 &  6.5 &  3.8 &  3.8 &  4.5 &  3.8 &  4.0 \\
\bottomrule
\end{tabular}
\caption{Accuracy and average sample budget for \textbf{DeepSeek R1}.}
\label{tab:deepseek_early_stopping_results}
\end{table*}

\clearpage

\subsection{Sample Failure Cases}
\label{sec:failure_cases}

In this section, we present sample cases from our  early-stopping method \cpearlyff where both CoT and PoT modalities reached early agreement on an incorrect answer. We also discuss the reasons for the incorrect agreement, and also provide the ground truth and the answer from full sampling--which is also most often wrong in such cases, as shown by our extremely high empirical probability of safety given agreement ($a_2$ in Table \ref{tab:parameter-probabilities}). 

\begin{figure*}[htb]
\centering
\begin{tcolorbox}[colback=white, colframe=black, title=Case 1: School Class Distribution (GPT-3.5 | GSM8K), left=1mm, fontupper=\footnotesize, breakable]
\textbf{Problem:} Each class in a school has 20 students. There are 3 classes. The school as a whole is 50\% boys and 50\% girls. The first class has 15 girls. The second class has 12 girls. How many boys are in the third class?

\textbf{Ground Truth Answer:} 17

\textbf{Full Sampling Answer (\cpfullmaj):} 3

\textbf{Early Stopping Answer (\cpearlyff):} 3

\textbf{CoT Solution:} Since each class has 20 students and there are 3 classes, there are a total of 20 × 3 = 60 students in the school. If the school as a whole is 50\% boys and 50\% girls, then there are an equal number of boys and girls in the school. This means that there are 30 boys and 30 girls in the school. The first class has 15 girls, so there are 20 - 15 = 5 boys in the first class. The second class has 12 girls, so there are 20 - 12 = 8 boys in the second class. In total, there are 30 - 15 - 12 = 3 boys in the third class. Therefore, there are $\boxed{3}$ boys in the third class.

\textbf{CoT Answer:} 3

\textbf{PoT Solution:}

\textbf{ Answer:} 3
\begin{verbatim}
total_students = 20 * 3
total_girls = 15 + 12
total_boys = total_students - total_girls
boys_in_third_class = total_students // 2 - total_girls
ans = boys_in_third_class
\end{verbatim}

\textbf{PoT Answer:} 3

\textbf{Discussion:} Both CoT and PoT make the same logical mistake about how to count the boys in the third class. They both incorrectly assume that if there are 30 total girls in the school and 27 girls in the first two classes (15 + 12), then the third class must have 3 boys. However, this confuses the number of remaining girls (30 - 27 = 3) with the number of boys in the third class. The correct solution requires first finding that the third class has 3 girls, and then calculating that it must have 20 - 3 = 17 boys to complete the class of 20 students. Both modalities reached early agreement because they encoded the same flawed reasoning, CoT through its step-by-step logic and PoT through the parallel calculation \texttt{total\_students // 2 - total\_girls}.
\end{tcolorbox}
\end{figure*}

\begin{figure*}[htb]
\centering
\begin{tcolorbox}[colback=white, colframe=black, title=Case 2: Animal Hunting Count (GPT-3.5 | GSM8K), left=1mm, fontupper=\footnotesize, breakable]
\textbf{Problem:} Last night Rick killed ten wolves and 15 cougars while hunting. Today Rick killed three times as many wolves as cougars and three fewer cougars than the previous night. How many animals did Rick kill?

\textbf{Ground Truth Answer:} 73

\textbf{Full Sampling Answer (\cpfullmaj):} 82

\textbf{Early Stopping Answer (\cpearlyff):} 82

\textbf{CoT Solution:} Let's denote the number of cougars Rick killed last night as x = 15. Since today Rick killed three fewer cougars than last night, he killed 15 - 3 = 12 cougars today. Today he killed three times as many wolves as cougars, so he killed 3 × 15 = 45 wolves today. The total number of wolves Rick killed is 10 (last night) + 45 (today) = 55 wolves. The total number of cougars Rick killed is 15 (last night) + 12 (today) = 27 cougars. Therefore, the total number of animals Rick killed is 55 + 27 = 82. Rick killed a total of $\boxed{82}$ animals.

\textbf{CoT Answer:} 82

\textbf{PoT Solution:}
\begin{verbatim}
wolves_last_night = 10
cougars_last_night = 15
wolves_today = 3 * cougars_last_night
cougars_today = cougars_last_night - 3
total_wolves = wolves_last_night + wolves_today
total_cougars = cougars_last_night + cougars_today
total_animals_killed = total_wolves + total_cougars
ans = total_animals_killed
\end{verbatim}

\textbf{PoT Answer:} 82

\textbf{Discussion:} Both CoT and PoT misinterpret the phrase ``three times as many wolves as cougars'' by using yesterday's cougar count (15) instead of today's cougar count (12) as the reference. The CoT reasoning explicitly states ``he killed 3 × 15 = 45 wolves today,'' while the PoT code encodes this same error with \texttt{wolves\_today = 3 * cougars\_last\_night}. The correct interpretation should compute today's wolves as 3 × 12 = 36, not 3 × 15 = 45, leading to a total of 10 + 36 + 15 + 12 = 73 animals instead of 82. This case demonstrates how ambiguous temporal references in natural language can lead both modalities to adopt the same incorrect semantic interpretation, resulting in cross-modal agreement on a wrong answer due to shared misunderstanding of which quantity serves as the multiplicative base.
\end{tcolorbox}
\end{figure*}

\begin{figure*}[htb]
\centering
\begin{tcolorbox}[colback=white, colframe=black, title=Case 3: Lemon Tree Investment (GPT-4o | GSM8K), left=1mm, fontupper=\footnotesize, breakable]
\textbf{Problem:} Carlos is planting a lemon tree. The tree will cost \$90 to plant. Each year it will grow 7 lemons, which he can sell for \$1.5 each. It costs \$3 a year to water and feed the tree. How many years will it take before he starts earning money on the lemon tree?

\textbf{Ground Truth Answer:} 13


\textbf{Full Sampling Answer (\cpfullmaj):} 12

\textbf{Early Stopping Answer (\cpearlyff):} 12

\textbf{CoT Solution:} To solve this problem, we need to determine how many years it will take for Carlos to recover his initial investment of \$90 and start earning a profit. Each year, the tree produces 7 lemons, and each lemon sells for \$1.50, so the annual revenue is 7 × 1.5 = 10.5 dollars. The annual cost of watering and feeding the tree is \$3. The annual net profit is 10.5 - 3 = 7.5 dollars. To recover the initial investment of \$90, we divide 90 by 7.5 to get 12 years. It will take Carlos 12 years to recover his initial investment. After 12 years, he will start earning money on the lemon tree. Final Answer: $\boxed{12}$

\textbf{CoT Answer:} 12

\textbf{PoT Solution:}
\begin{verbatim}
# Define the given values 
planting_cost = 90 
lemons_per_year = 7 
price_per_lemon = 1.5 
annual_cost = 3 
# Calculate the annual profit 
annual_revenue = lemons_per_year * price_per_lemon 
annual_profit = annual_revenue - annual_cost 
# Calculate the number of years required to break even 
years = 0 
total_profit = 0 
while total_profit < planting_cost: 
    years += 1 
    total_profit += annual_profit 
ans = years
\end{verbatim}

\textbf{PoT Answer:} 12

\textbf{Discussion:} Both CoT and PoT make a boundary condition error by confusing the year Carlos breaks even with the year he starts earning money. The CoT reasoning correctly calculates that 12 years of \$7.5 annual profit equals \$90, but concludes ``After 12 years, he will start earning money,'' when in fact year 12 is when he breaks even. The PoT solution implements a while loop with condition \texttt{total\_profit < planting\_cost}, which exits when the cumulative profit reaches \$90, returning 12 as the break-even year rather than 13 as the first profitable year. The correct answer requires recognizing that after 12 years Carlos has exactly \$90 in cumulative profit (breaking even), and only in year 13 does he start earning money beyond his initial investment. This off-by-one error demonstrates how both modalities can encode the same conceptual confusion between ``breaking even'' and ``starting to earn money.''
\end{tcolorbox}
\end{figure*}

\begin{figure*}[htb]
\centering
\begin{tcolorbox}[colback=white, colframe=black, title=Case 4: Salary Increase Policy (GPT-4o | GSM8K), left=1mm, fontupper=\footnotesize, breakable]
\textbf{Problem:} A company pays each of its employees \$600 in a month. The company has a policy of increasing the salaries of each of its employees by 10\% of the initial salary every year for those who've stayed in the company for five years. If Sylvie just clocked 5 years in the company last December, what's her annual salary after three more years of service?

\textbf{Ground Truth Answer:} 9360


\textbf{Full Sampling Answer (\cpfullmaj):} 9360

\textbf{Early Stopping Answer (\cpearlyff):} 7920

\textbf{CoT Solution:} We are tasked with finding Sylvie's annual salary after three more years of service. Sylvie's initial monthly salary is \$600. After 5 years, her salary increases by 10\% of the initial salary. We need to calculate her annual salary after 3 more years of service (8 years total). The increase is 10\% × 600 = 60 dollars, so after 5 years, her new monthly salary becomes 600 + 60 = 660 dollars. Since the problem does not mention any further increases, her monthly salary remains constant at 660 dollars for the next 3 years. To find her annual salary after 8 years, we multiply her monthly salary by 12: 660 × 12 = 7,920 dollars. Sylvie's annual salary after 8 years of service is $\boxed{7920 \text{ dollars}}$

\textbf{CoT Answer:} 7920 dollars

\textbf{PoT Solution:}
\begin{verbatim}
initial_monthly_salary = 600
annual_salary = initial_monthly_salary * 12
increment_rate = 0.10

# Calculate incremented salary after 5 years
incremented_salary = annual_salary * (1 + increment_rate)

# Calculate annual salary after three more years
# Since the salary doesn't increase further after 5 years, it remains the same
ans = incremented_salary
\end{verbatim}

\textbf{PoT Answer:} 7920.0

\textbf{Discussion:} Both CoT and PoT misinterpret the salary increase policy by treating it as a one-time adjustment rather than an annual recurring increase. The CoT reasoning explicitly states ``Since the problem does not mention any further increases, her monthly salary remains constant at 660 dollars for the next 3 years,'' applying only a single 10\% increase at year 5. The PoT solution mirrors this interpretation with the comment ``Since the salary doesn't increase further after 5 years, it remains the same'' and calculates only \texttt{annual\_salary * (1 + increment\_rate)} once. The correct interpretation is that the policy grants a 10\% raise every year after the initial 5-year milestone, so Sylvie should receive increases at years 6, 7, and 8, yielding an annual salary of \$7200 × $(1.1)^3$ = \$9,360. This case illustrates how ambiguous wording about recurring versus one-time benefits can lead both modalities to encode the same oversimplified policy model, resulting in cross-modal agreement on an incorrect answer.
\end{tcolorbox}
\end{figure*}

\begin{figure*}[htb]
\centering
\begin{tcolorbox}[colback=white, colframe=black, title=Case 5: Stamp Price Doubling (Qwen-Coder | MATH), left=1mm, fontupper=\footnotesize, breakable]
\textbf{Problem:} Mike paid \$1.25 for a stamp three years ago. He was just offered double that amount for the stamp. Assuming the stamp's offer price doubles every three years, how many dollars will he be offered in 12 more years?

\textbf{Ground Truth Answer:} \$40


\textbf{Full Sampling Answer (\cpfullmaj):} 20

\textbf{Early Stopping Answer (\cpearlyff):} 20

\textbf{CoT Solution:} We are given that Mike paid \$1.25 for a stamp three years ago, and the offer price doubles every three years. We need to find how much he will be offered in 12 more years. The price doubles every 3 years. Starting from the initial price of \$1.25: After 3 years: \$1.25 × 2 = \$2.50. After 6 years: \$2.50 × 2 = \$5.00. After 9 years: \$5.00 × 2 = \$10.00. After 12 years: \$10.00 × 2 = \$20.00. In 12 years, there are 12 ÷ 3 = 4 doubling periods, so the price will have doubled 4 times. The final price is \$1.25 × $2^4$ = \$1.25 × 16 = \$20. Therefore, Mike will be offered $\boxed{20}$ dollars in 12 more years.

\textbf{CoT Answer:} 20

\textbf{PoT Solution:}
\begin{verbatim}
initial_price = 1.25
years = 12
doubling_period = 3
num_doublings = years // doubling_period
future_price = initial_price * (2 ** num_doublings)
ans = future_price
\end{verbatim}

\textbf{PoT Answer:} 20.0

\textbf{Discussion:} Both CoT and PoT fail to recognize that the current offer of \$2.50 (double the original \$1.25) serves as the starting point for future doublings, not the original purchase price. The CoT reasoning begins with ``Starting from the initial price of \$1.25'' and applies 4 doublings over 12 years, while the PoT solution uses \texttt{initial\_price = 1.25} as the base value. The correct interpretation is that Mike currently has an offer of \$2.50, and in 12 more years (4 additional doubling periods), the price will be \$2.50 × $2^4$ = \$2.50 × 16 = \$40. This temporal reference error shows how both modalities can misidentify the proper baseline when a problem describes both historical and future time periods, leading to agreement on an answer that's off by a factor of 2 due to the shared misinterpretation of which price point represents ``now.''
\end{tcolorbox}
\end{figure*}

\begin{figure*}[htb]
\centering
\begin{tcolorbox}[colback=white, colframe=black, title=Case 6: Bus Schedule Reading (Qwen-Coder | TabMWP), left=1mm, fontupper=\footnotesize, breakable]
\textbf{Problem:} Look at the following schedule. Sharon got on the bus at downtown at 11:45 A.M. What time will she get to the art museum? 

Tour bus schedule:
\begin{center}
\begin{tabular}{lccc}
Location & Time 1 & Time 2 & Time 3 \\
\hline
skyscraper & 9:30 A.M. & 10:15 A.M. & 10:45 A.M. \\
downtown & 10:30 A.M. & 11:15 A.M. & 11:45 A.M. \\
city hall & 11:15 A.M. & 12:00 P.M. & 12:30 P.M. \\
art museum & 12:15 P.M. & 1:00 P.M. & 1:30 P.M. \\
the zoo & 1:15 P.M. & 2:00 P.M. & 2:30 P.M.
\end{tabular}
\end{center}

\textbf{Ground Truth Answer:} 1:30 P.M.


\textbf{Full Sampling Answer (\cpfullmaj):} 12:15 \text{ P.M.}

\textbf{Early Stopping Answer (\cpearlyff):} 12:15 \text{ P.M.}

\textbf{CoT Solution:} Looking at the bus schedule, I need to find when Sharon will reach the art museum after getting on at downtown at 11:45 A.M. Sharon gets on the bus at downtown at 11:45 A.M. Looking at the downtown row, I see times: 10:30 A.M., 11:15 A.M., 11:45 A.M. Since Sharon gets on at 11:45 A.M. at downtown, I need to find what time the art museum stop is for that same bus route. Looking at the art museum row, I see times: 12:15 P.M., 1:00 P.M., 1:30 P.M. The 11:45 A.M. downtown departure corresponds to the 12:15 P.M. art museum arrival time. Therefore, Sharon will get to the art museum at $\boxed{12:15 \text{ P.M.}}$

\textbf{CoT Answer:} 12:15 P.M.

\textbf{PoT Solution:}
\begin{verbatim}
# Looking at the schedule table
# Downtown: 11:45 A.M. (Sharon gets on the bus)
# Art museum: 12:15 P.M. (Sharon arrives)
ans = "12:15 P.M."
\end{verbatim}

\textbf{PoT Answer:} 12:15 P.M.

\textbf{Discussion:} Both CoT and PoT incorrectly align the table columns by assuming the 11:45 A.M. downtown time corresponds to Time 1 rather than Time 3. The CoT reasoning states ``The 11:45 A.M. downtown departure corresponds to the 12:15 P.M. art museum arrival time,'' incorrectly pairing the third column entry in the downtown row with the first column entry in the art museum row. The PoT solution directly hardcodes this same misalignment with the comment mapping downtown 11:45 A.M. to art museum 12:15 P.M. The correct interpretation requires recognizing that 11:45 A.M. appears in the Time 3 column, so Sharon should arrive at the art museum at the Time 3 entry of 1:30 P.M. This demonstrates how both modalities can make identical table-reading errors when they fail to track column alignment, leading to cross-modal agreement on an answer based on a shared structural misinterpretation of tabular data.
\end{tcolorbox}
\end{figure*}

\begin{figure*}[htb]
\centering
\begin{tcolorbox}[colback=white, colframe=black, title=Case 7: Reality Show Probability (Mistral-Large | TabMWP), left=1mm, fontupper=\footnotesize, breakable]
\textbf{Problem:} In the first episode of a reality show, contestants had to spin two wheels of fate. Spinning the first wheel determined the remote location where contestants would reside. Spinning the second wheel determined which "bonus survival tool" they would be allowed to bring. What is the probability that a randomly selected participant spun the first wheel and landed on desert and spun the second wheel and landed on a tent? Simplify any fractions.

\begin{center}
\begin{tabular}{lcc}
          & A tent & Matches \\
\hline
Desert    & 6      & 7       \\
Rainforest & 2     & 5      
\end{tabular}
\end{center}

\textbf{Ground Truth Answer:} 3/10


\textbf{Full Sampling Answer (\cpfullmaj):} 13/50

\textbf{Early Stopping Answer (\cpearlyff):} 13/50

\textbf{CoT Solution:} To solve this problem, we need to calculate the probability of two independent events happening together: landing on "desert" with the first wheel and landing on "a tent" with the second wheel. For the first wheel (location): Desert has 6 + 7 = 13 outcomes, Rainforest has 2 + 5 = 7 outcomes, for a total of 20 outcomes. For the second wheel (survival tool): A tent has 6 + 2 = 8 outcomes, Matches has 7 + 5 = 12 outcomes, for a total of 20 outcomes. The probability of landing on "desert" is 13/20. The probability of landing on "a tent" is 8/20 = 2/5. Since the events are independent, the combined probability is: P(desert and a tent) = 13/20 × 2/5 = 26/100 = 13/50. The final answer is $\boxed{\frac{13}{50}}$

\textbf{CoT Answer:} 13/50

\textbf{PoT Solution:}
\begin{verbatim}
# Input parameters
desert_tent = 6
desert_matches = 7
rainforest_tent = 2
rainforest_matches = 5

# Total outcomes for the first wheel
total_first_wheel = desert_tent + desert_matches + rainforest_tent + rainforest_matches

# Total outcomes for the second wheel
total_second_wheel = desert_tent + desert_matches + rainforest_tent + rainforest_matches

# Probability of landing on Desert with the first wheel
prob_desert = (desert_tent + desert_matches) / total_first_wheel

# Probability of landing on A tent with the second wheel
prob_tent = (desert_tent + rainforest_tent) / total_second_wheel

# Probability of both events happening together
ans = prob_desert * prob_tent

# Simplify the fraction
from fractions import Fraction
ans = Fraction(ans).limit_denominator()
\end{verbatim}

\textbf{PoT Answer:} 13/50

\textbf{Discussion:} Both CoT and PoT misinterpret the table structure by treating the two wheels as independent when the table actually represents a joint probability distribution where the cell values are frequencies of combined outcomes. The CoT reasoning incorrectly calculates marginal probabilities by summing rows and columns (P(desert) = 13/20, P(tent) = 8/20) and then multiplying them as if the wheels were independent. The PoT solution encodes this same error by computing \texttt{prob\_desert} and \texttt{prob\_tent} separately from marginal totals and multiplying them together. The correct interpretation is that the table directly shows 6 participants landed on both desert and tent out of a total of 6 + 7 + 2 + 5 = 20 participants, yielding P(desert and tent) = 6/20 = 3/10. This case demonstrates how both modalities can misunderstand the semantic meaning of tabular data, treating a contingency table as if it described independent marginal distributions rather than joint outcomes.
\end{tcolorbox}
\end{figure*}

\begin{figure*}[htb]
\centering
\begin{tcolorbox}[colback=white, colframe=black, title=Case 8: Modular Arithmetic with Digit Composition (GPT-4o | MATH), left=1mm, fontupper=\footnotesize, breakable]
\textbf{Problem:} Suppose that a 30-digit integer $N$ is composed of thirteen 7s and seventeen 3s. What is the remainder when $N$ is divided by 36?

\textbf{Ground Truth Answer:} 25


\textbf{Full Sampling Answer (\cpfullmaj):} 34

\textbf{Early Stopping Answer (\cpearlyff):} 34

\textbf{CoT Solution:} To find the remainder when $N$ is divided by 36, we use the Chinese Remainder Theorem since $36 = 4 \times 9$. We find $N \bmod 4$ and $N \bmod 9$ separately. For $N \bmod 4$: Since $7 \equiv 3 \pmod{4}$ and $3 \equiv 3 \pmod{4}$, every digit contributes 3 modulo 4. The sum of digits is $13(7) + 17(3) = 142$. Thus $N \equiv 142 \equiv 2 \pmod{4}$. For $N \bmod 9$: The sum of digits is 142, and $1+4+2 = 7$, so $N \equiv 7 \pmod{9}$. Using CRT, we solve: $N \equiv 2 \pmod{4}$ and $N \equiv 7 \pmod{9}$. Let $N = 4k + 2$. Then $4k + 2 \equiv 7 \pmod{9}$, so $4k \equiv 5 \pmod{9}$. The multiplicative inverse of 4 mod 9 is 7, so $k \equiv 35 \equiv 8 \pmod{9}$. Thus $N = 4(9m + 8) + 2 = 36m + 34$. Therefore $N \equiv \boxed{34} \pmod{36}$

\textbf{CoT Answer:} 34

\textbf{PoT Solution:}
\begin{verbatim}
from sympy import mod_inverse

num_sevens = 13
num_threes = 17

# Calculate N modulo 4 using digit sum
remainder_mod_4 = (num_sevens * 7 + num_threes * 3) % 4

# Calculate N modulo 9 (sum of digits)
sum_of_digits = num_sevens * 7 + num_threes * 3
remainder_mod_9 = sum_of_digits % 9

# Chinese Remainder Theorem
a1, m1 = remainder_mod_4, 4
a2, m2 = remainder_mod_9, 9

m1_inv = mod_inverse(m1, m2)
m2_inv = mod_inverse(m2, m1)

x = (a1 * m2 * m2_inv + a2 * m1 * m1_inv) % (m1 * m2)

ans = x
\end{verbatim}

\textbf{PoT Answer:} 34

\textbf{Discussion:} Both CoT and PoT incorrectly compute $N \bmod 4$ by using the digit sum rather than considering positional values. The CoT reasoning states ``every digit contributes 3 modulo 4'' and calculates the sum of digits (142) mod 4, while the PoT solution implements \texttt{remainder\_mod\_4 = (num\_sevens * 7 + num\_threes * 3) \% 4}, both treating $N$ as if it equals the sum of its digits modulo 4. However, for modulo 4, the last two digits determine the remainder, not the digit sum. The correct approach recognizes that the actual value of $N$ depends on digit positions: with thirteen 7s and seventeen 3s, the last two digits could be 77, 73, 37, or 33, yielding $N \equiv 1, 1, 1, \text{ or } 1 \pmod{4}$ respectively. Combined with $N \equiv 7 \pmod{9}$, the Chinese Remainder Theorem gives $N \equiv 25 \pmod{36}$. This case illustrates how both modalities can adopt an oversimplified mathematical abstraction that works for modulo 9 but incorrectly generalizes to modulo 4.
\end{tcolorbox}
\end{figure*}